\documentclass[11pt]{article}

\usepackage[final]{acl}

\usepackage{times}
\usepackage{latexsym}
\usepackage{mathtools}

\usepackage[T1]{fontenc}

\usepackage[utf8]{inputenc}

\usepackage{microtype}

\usepackage{inconsolata}

\usepackage{graphicx}
\usepackage{amsmath}

\usepackage{booktabs}
\usepackage[table,dvipsnames]{xcolor}
\usepackage{amssymb}
\usepackage{xspace}
\usepackage{enumitem}
\usepackage{threeparttable}
\usepackage{wrapfig}
\usepackage{multirow}
\usepackage{pifont}
\usepackage{stackengine}    
\newcommand{\cmark}{\ding{51}}
\newcommand{\xmark}{\ding{55}}

\newcommand{\Passk}{Pass\textasciicircum{}k}

\usepackage{array}
\usepackage{tabularx}
\usepackage{stfloats}
\usepackage{ragged2e}

\providecommand{\citep}[1]{\cite{#1}}
\providecommand{\citet}[1]{\cite{#1}}

\definecolor{turnsblue}{RGB}{98, 122, 156}
\definecolor{pipelineblue}{RGB}{143, 158, 201}
\definecolor{pipelinepeach}{RGB}{218, 174, 149}
\definecolor{pipelinepurple}{RGB}{188, 169, 201}
\definecolor{pipelineloop}{RGB}{192, 57, 43}
\definecolor{tablegroupbg}{RGB}{232, 232, 250}
\definecolor{casebeige}{RGB}{246, 238, 226}
\newcommand{\sv}[2]{%
  \begingroup
  \pgfmathparse{#1/100}%
  \pgfmathprintnumber[fixed,precision=2]{\pgfmathresult}%
  \endgroup
}

\usepackage{tikz}
\usetikzlibrary{arrows.meta,positioning,shapes.geometric,fit,backgrounds,calc}

\usepackage{tcolorbox}
\usepackage{mdframed}
\tcbuselibrary{listingsutf8,breakable,skins}
\tcbset{%
  coltext=black,
  after={\color{black}}
}
\usepackage{listings}
\tcbuselibrary{breakable,skins}
\usepackage{tabularx}
\newcolumntype{Y}{>{\RaggedRight\arraybackslash}X}
\newcommand{\codefmt}[1]{{\ttfamily\detokenize{#1}}}

\newcommand{\model}[1]{#1}
\newcommand{\tabgroup}[1]{%
  \addlinespace[0.8ex]
  \rowcolor{tablegroupbg}[0pt][0pt]
  \multicolumn{14}{@{}c@{}}{\rule{0pt}{2.0ex}\normalsize\bfseries\itshape #1}\\[-0.25ex]
  \arrayrulecolor{black}
}

\title{EHR-Complex: Benchmarking Medical Agents for Complex Clinical Reasoning}

\author{
\textbf{Yitong Qiao\textsuperscript{1}}\thanks{Equal contribution.}\thanks{Work done during an internship at Ant Group.},
\textbf{Lei Liu\textsuperscript{1,2}}\footnotemark[1],
\textbf{Yue Shen\textsuperscript{2}},
\textbf{Jian Wang\textsuperscript{2}},
\textbf{Jinjie Gu\textsuperscript{2}},
\textbf{Zhixuan Chu\textsuperscript{1}}
\textbf{Kui Ren\textsuperscript{1}}
\\
\\
\textsuperscript{1}Zhejiang University \quad
\textsuperscript{2}Ant Group
\\
\small{
\textbf{Correspondence:} \href{mailto:zhixuanchu@zju.edu.cn}{zhixuanchu@zju.edu.cn}
}
}

\begin{document}
\maketitle
\begin{abstract}
Clinical agents promise to democratize access to electronic health records (EHRs), yet existing benchmarks fail to reflect the complexity of practical EHR analysis, \textit{e.g.}, often operating on idealized, clean EHRs via static SQL generation rather than interactive execution. In this work, we introduce \textbf{EHR-Complex}, a large-scale benchmark designed for interactive clinical database reasoning. Built on the large MIMIC-IV substrate (365K patients, 31 tables, 500M+ records), EHR-Complex comprises about 52K tasks spanning six clinical intents, supporting both patient-level and population-level queries, where each task requires an agent to interact with a sandboxed environment by executing SQL queries or Python code. Notably, EHR-Complex considers the real-world SQL task complexity for longitudinal multi-table aggregation and compositional reasoning, resulting in 31.93 SQL structural components per query on average. Evaluation results on EHR-Complex reveal the clinical difficulty of these EHR reasoning scenarios, with the top-performing model achieving only 62.3\% exact-match accuracy. \Passk{} consistency drops below 50\% for nearly all evaluated models at k$=4$, exposing broad stochastic fragility. A fine-grained analysis of more than 3,800 failed trajectories for representative LLMs reveals three dominant failure modes: SQL logic errors, medical-code lookup failures, and semantic misunderstandings. EHR-Complex provides a rigorous testbed for clinical agents and highlights remaining gaps in robust reasoning for large-scale EHR analysis.
\end{abstract}

\section{Introduction}

Large language models (LLMs) and agents are increasingly viewed as promising interfaces for democratizing access to electronic health records (EHRs), potentially enabling clinicians and healthcare administrators to retrieve and analyze patient data through natural language rather than specialized database expertise~\citep{wang2020text,lee_ehrsql_2023,xu_medagentgym_2025,jiang2025medagentbench,lee_fhir-agentbench_2025}. Such capability is especially valuable in modern clinical environments, where EHR databases are large, heterogeneous, longitudinal, and difficult to query efficiently. Clinical agents could substantially reduce the technical barrier to complex cohort construction, patient-level record retrieval, and retrospective outcome analysis.

Prior work on executable agent benchmarks, such as SWE-bench~\citep{jimenez_swebench_2023}, WebArena~\citep{zhou_webarena_2023}, and MLAgentBench~\citep{huang2024mlagentbench}, has shown that grounding agents in environments with execution feedback can expose failure modes that static evaluation may miss. Despite this promise, current evaluation settings for clinical database reasoning still only partially capture the complexity of practical EHR analysis. However, existing benchmarks often simplify the problem into one-shot text-to-SQL generation over clean, idealized schemas, where the model is evaluated only on a final static query string. 

Practical EHR analysis often requires agents to operate in interactive environments rather than merely produce syntactically valid single-shot SQL statements: users must inspect intermediate results, resolve schema ambiguities, identify relevant medical codes, perform longitudinal aggregation across multiple tables, and iteratively refine logic when initial queries fail or return unexpected outputs.
Moreover, large EHR databases are noisy, expansive, and semantically challenging, requiring models to reason jointly over temporal structure, coding systems, and compositional constraints.

\begin{figure*}[t]
  \centering
  \begin{minipage}[b]{0.50\textwidth}
    \centering
    \includegraphics[width=\linewidth]{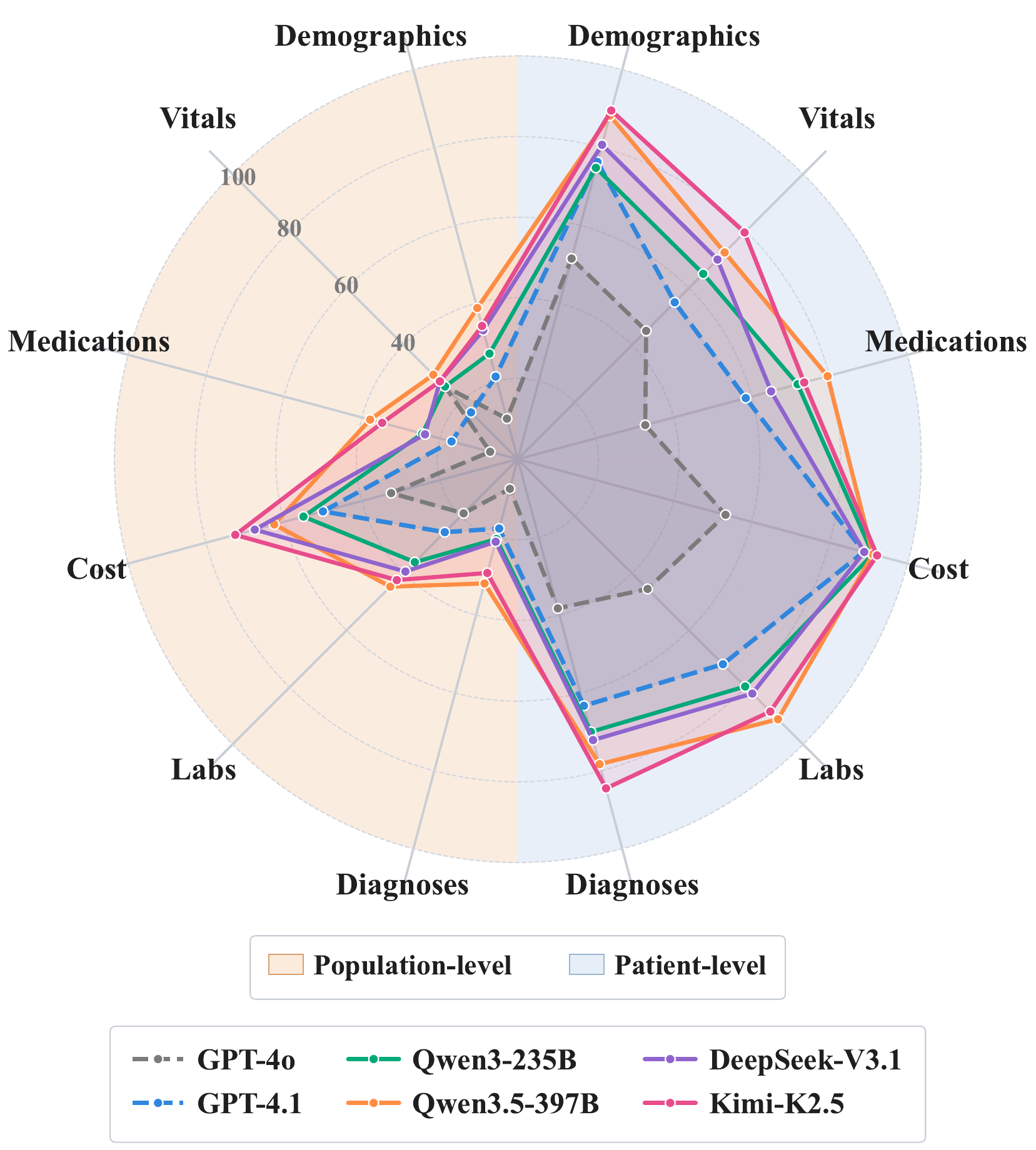}
    \par\vspace{0.2ex}
    {\small\textbf{(a)}~Success rates across 12 (6 intent\,×\,2 scope) dimensions.}%
  \end{minipage}%
  \hfill
  \begin{minipage}[b]{0.50\textwidth}
    \centering
    \includegraphics[width=\linewidth]{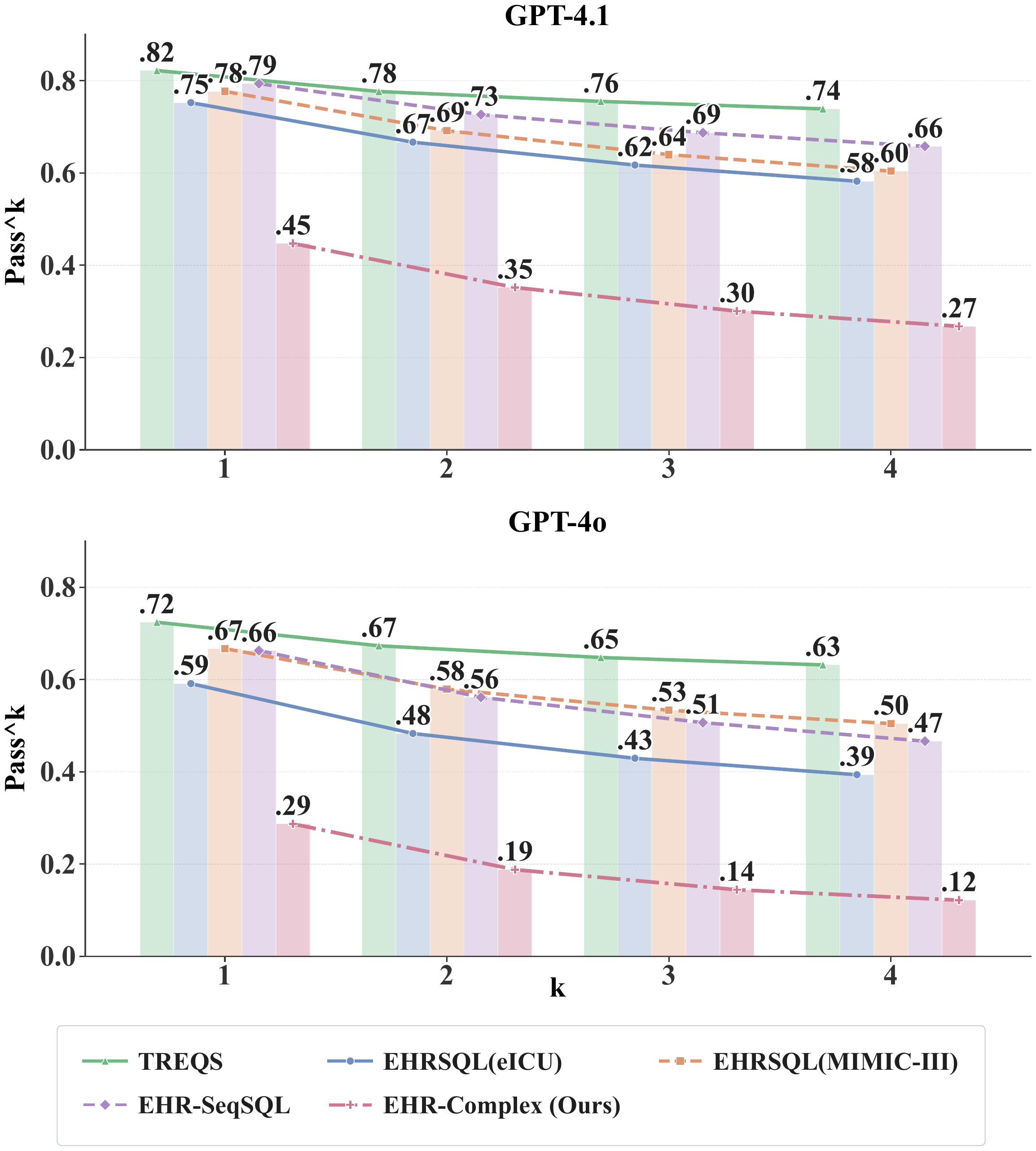}
    \par\vspace{0.2ex}
    {\small\textbf{(b)}~\Passk{} across different benchmarks.}
  \end{minipage}
  \vspace{-0.5ex}
  \caption{%
    \textbf{Overall Performance of LLMs on EHR-Complex Test Set.}
    \textbf{(a)}~Performance across 12 (6 intent\,×\,2 scope) dimensions. Patient-level queries typically follow one clinical evidence path, whereas population-level queries require aligning and aggregating evidence paths across patients; any misaligned path may lead to an incorrect result.
    \textbf{(b)}~Evaluated models achieve near-perfect \Passk{} on legacy benchmarks but struggle on EHR-Complex.
  }
  \label{fig:teaser}
\end{figure*}

To this end, we introduce \textbf{EHR-Complex}, a large-scale benchmark for interactive clinical database reasoning.
Built on the full MIMIC-IV substrate, containing \textbf{365K} patients, \textbf{31} tables, and more than \textbf{500 million} records, EHR-Complex consists of approximately \textbf{52K} tasks spanning six major clinical intents.

The benchmark supports both patient-level and population-level information needs and requires agents to operate in a sandboxed environment by executing SQL queries or Python code, rather than merely generating static answers.
Importantly, EHR-Complex is designed to reflect the compositional and longitudinal nature of large-scale EHR analysis: benchmark tasks require substantial multi-table reasoning and aggregation, with queries averaging 31.93 SQL structural components per query. Our empirical evaluation across a range of state-of-the-art LLMs shows that EHR-Complex remains highly challenging.
Even the best-performing model achieves 62.3\% exact-match accuracy, indicating a substantial gap between current capabilities and the robustness needed for dependable EHR analysis. Overall contributions are as follows:

\begin{itemize}[leftmargin=*]
    \item We introduce a large-scale EHR-Complex benchmark for interactive clinical database reasoning. Built on the full MIMIC-IV substrate, EHR-Complex comprises approximately 52K tasks across six clinical intents, covering both patient-level and population-level scopes.

    \item We propose to incorporate real-world SQL task complexity into EHR-Complex, \textit{e.g.}, longitudinal multi-table aggregation and compositional reasoning. Each task is paired with an execution-validated SQL program and a verified answer, with queries averaging 31.93 SQL structural components per query.

    \item We evaluate 12 LLM agents and analyze their robustness and failure modes, showing persistent challenges in population-level reasoning, medical-code grounding, and SQL logic; we also report a controlled in-domain SFT validation using successful trajectories.
\end{itemize}

\begin{figure*}[ht]
\centering
\includegraphics[width=0.98\textwidth]{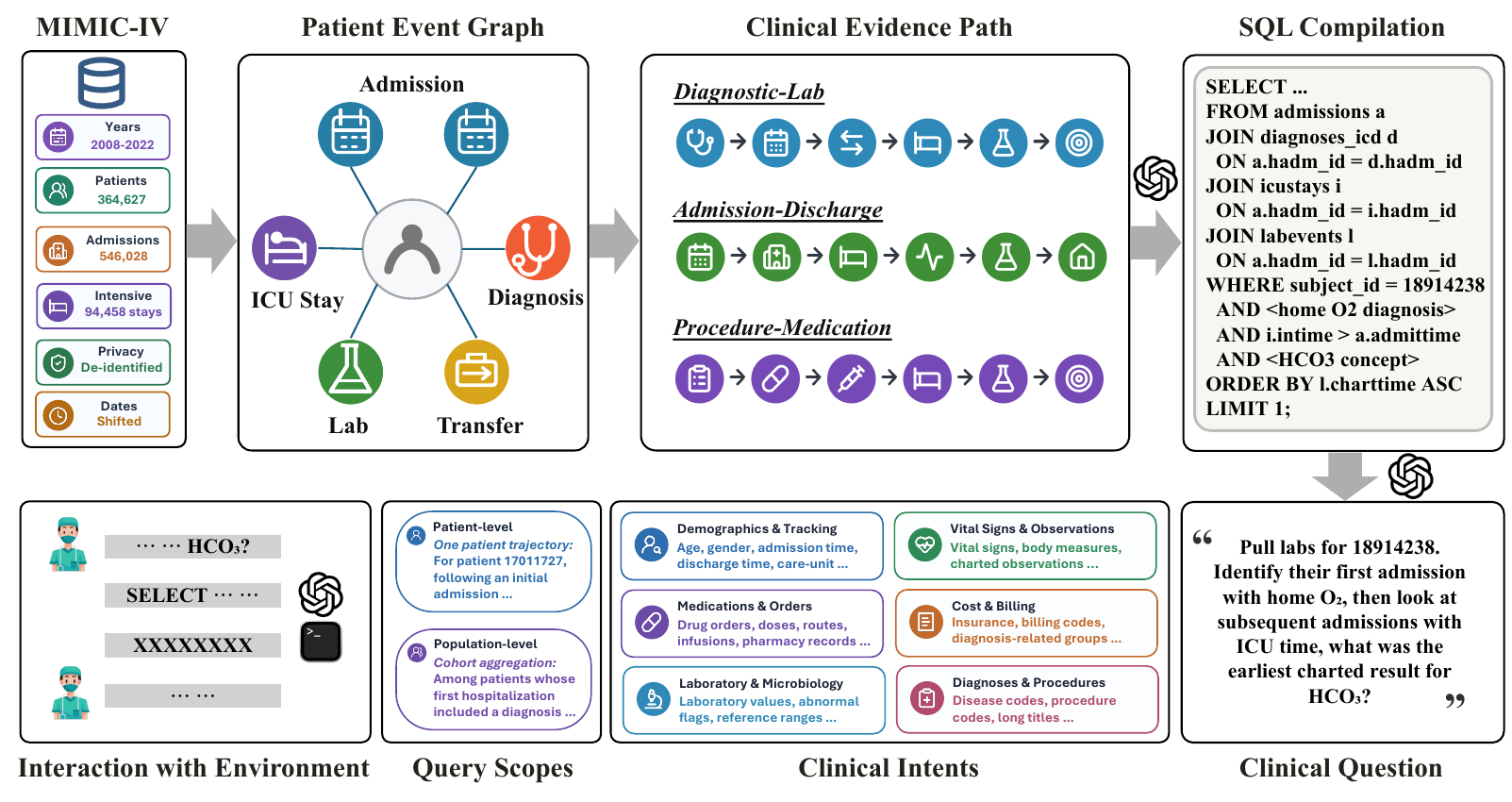}
\caption{EHR-Complex Construction Pipeline. \textbf{Step-1:} Based on the full MIMIC-IV substrate, patient event graphs are derived for the subsequent compositional reasoning. \textbf{Step-2:} Clinical evidence paths are extracted from event graphs as SQL query candidates. \textbf{Step-3:} Clinical evidence paths are compiled into execution-validated SQL templates. These templates are materialized into patient- and population-level questions across six clinical intents. EHR-Complex provides the environment sandboxes for interactive SQL/Python execution.
}
\label{fig:data_pipeline}
\end{figure*}

\section{Related Work}

\paragraph{Static Text-to-SQL.}
Early efforts to enable natural-language access to EHRs primarily focused on semantic parsing, inheriting the cross-schema evaluation established by general-domain benchmarks like Spider~\citep{yu2018spider} while adapting it for domain-specific medical grounding and temporal logic. Foundational clinical datasets such as MIMICSQL~\citep{wang2020text} and emrKBQA~\citep{raghavan2021emrkbqa} mapped medical concepts to schemas, while later works like EHRSQL~\citep{lee_ehrsql_2023} and EHR-SeqSQL~\citep{ryu2024ehr} introduced hospital-derived questions with time-sensitive templates. Although these resources advanced clinically grounded text-to-SQL evaluation, they remain constrained by a static generation paradigm that lacks execution feedback. Furthermore, to ensure tractability, they typically rely on heavily filtered patient subsets ($<$1K) and fewer schemas (5-17 tables).

\paragraph{Medical Agent Benchmarks.}

Inspired by general-domain interactive benchmarks like SWE-bench~\citep{jimenez_swebench_2023}, WebArena~\citep{zhou_webarena_2023}, and MLAgentBench~\citep{huang2024mlagentbench}, recent efforts have reframed EHR access as a multi-turn agentic task. Frameworks such as EHRAgent~\citep{shi_ehragent_2024}, FHIR-AgentBench~\citep{lee_fhir-agentbench_2025}, MedAgentGym~\citep{xu_medagentgym_2025}, and MedAgentBench~\citep{jiang2025medagentbench} now provide executable environments for iterative querying, marking a pivotal shift from static generation to interactive problem-solving. While related non-medical benchmarks (\textit{e.g.}, $\tau$-bench~\citep{yao2024tau}, WorkArena~\citep{drouin2024workarena}) and quantitative medical tasks like MedCalc-Bench~\citep{khandekar2024medcalc} similarly emphasize grounded interaction, a critical limitation persists: existing interactive medical benchmarks largely adapt task instances from legacy static datasets (\textit{e.g.}, EHRSQL, TREQS) rather than designing native challenges for complex, large-scale agent reasoning.

Overall, existing EHR benchmarks are ill-suited for the complexity of real-world EHR analysis, \textit{e.g.}, execution-guided schema exploration, longitudinal multi-table aggregation, and population-level cohort reasoning over large-scale databases.

\begin{table*}[htbp]
    \centering
    \caption{
    Comparison with Existing EHR Benchmarks. EHR-Complex reflects real-world clinical complexity, \textit{e.g.}, longitudinal multi-table aggregation and compositional reasoning.
    }
    \label{tab:comparison}
    \renewcommand\arraystretch{1.1}
    \resizebox{\linewidth}{!}{ %
    \begin{tabular}{l|c|ccc|ccc|c}
    \toprule
    \multirow{2}{*}{\textbf{Dataset}} & \textbf{Agentic} & \multicolumn{3}{c|}{\textbf{Database Scale}} & \multicolumn{3}{c|}{\textbf{Task Instances}} & \textbf{Task} \\
    \cmidrule(lr){3-5} \cmidrule(lr){6-8}
    & \textbf{Mode$^{2}$} & \textbf{\#Patients} & \textbf{\#Tables} & \textbf{\#Elements} & \textbf{Category} & \textbf{\#Test} & \textbf{\#Train} & \textbf{Complexity}$^{3}$\\
    \midrule
    MedAgentBench$^{1}$~\citep{jiang2025medagentbench} & \cmark & 100 & -- & 700K & 10 & 300 & -- & -- \\
    TREQS~\citep{wang2020text} & \xmark & 100 & 5 & 2.5M & 4 & 996 & 8,988 & 3.06 \\
    EHRSQL (eICU)~\citep{lee_ehrsql_2023} & \xmark & <1K & 10 & 1.5M & 9 & 611 & 6,213 & 11.21 \\
    EHRSQL (MIMIC-III)~\citep{lee_ehrsql_2023} & \xmark & <1K & 17 & 1.4M & 9 & 1,122 & 9,318 & 11.30 \\
    EHR-SeqSQL~\citep{ryu2024ehr} & \xmark & <1K & 17 & 1.4M & 4 & 7,913 & 18,950  & 17.03 \\
    \midrule
    \rowcolor{RoyalPurple!6} \textbf{EHR-Complex (Ours)} & \textbf{\cmark} & \textbf{365K} & \textbf{31} & \textbf{>500M} & \textbf{12} &  \textbf{3,915} & \textbf{48,092} & \textbf{31.93} \\
    \bottomrule
    \end{tabular}
    } 
    \vspace{0.5ex} 
    \begin{minipage}{\linewidth}
        \small
        \begin{itemize}[leftmargin=1.5em, labelsep=0.5em, itemsep=0pt, parsep=0pt]
            \item[$^{1}$] MedAgentBench operates in an API environment, making it incompatible with SQL complexity metrics.
            \item[$^{2}$] Agentic mode indicates whether a benchmark supports executable interaction with an EHR environment.
            \item[$^{3}$] Average total number of SQL structural components per query. Figure~\ref{fig:sql_complexity_dimensions} shows the breakdown.
        \end{itemize}
    \end{minipage}
\end{table*}

\begin{figure*}[t]
  \centering
  \includegraphics[width=\textwidth]{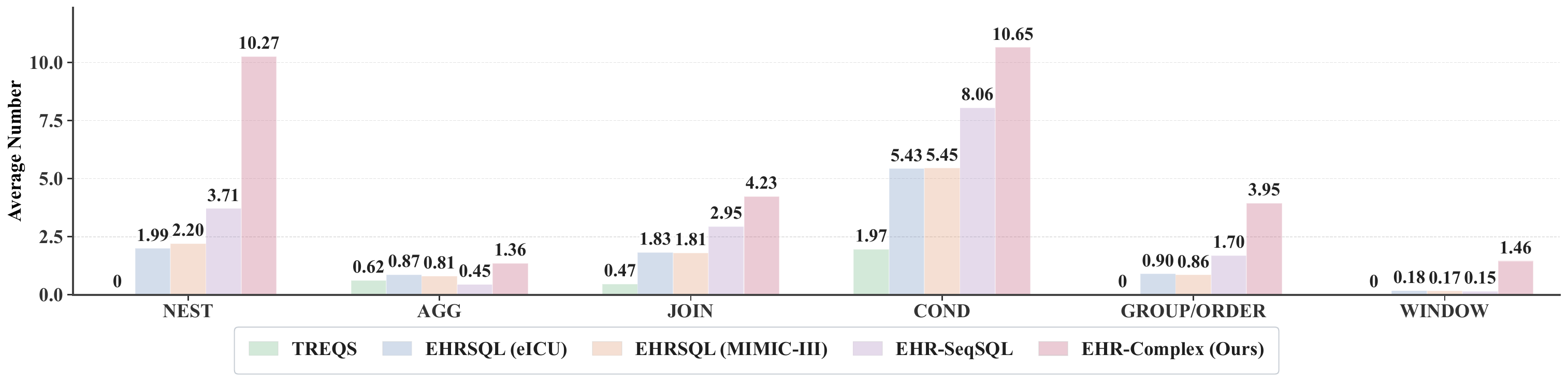}
  \caption{Average Number of SQL Structural Components Per Query Across Different Benchmarks. EHR-Complex exhibits higher task complexity for multi-table aggregation and compositional reasoning.}
  \label{fig:sql_complexity_dimensions}
\end{figure*}

\section{EHR-Complex Benchmark}

\paragraph{Task Formulation.} 
As illustrated in Figure~\ref{fig:data_pipeline}, given a natural-language query $x \in \mathcal{X}$ over a database $\mathcal{D}$ with schema $\Sigma$, an agent produces a multi-turn trajectory $\tau \in \mathcal{T}$ by executing SQL queries or Python code and observing environment feedback. The objective is to derive the target output $y \in \mathcal{Y}$ that matches the verified answer $y^*$.

For each trajectory, the correctness function is formulated as $\mathcal{E}: \mathcal{T} \times \mathcal{Y} \rightarrow \{0,1\}$:
\begin{equation}
    \mathcal{E}(\tau, y^*) = \mathbb{I}(y = y^*),
\end{equation}
where $y$ denotes the final answer extracted from trajectory $\tau$, $y^*$ denotes the ground-truth output, and $\mathbb{I}(\cdot)$ is the indicator function.

Unlike existing EHR benchmarks~\citep{wang2020text,lee_ehrsql_2023,ryu2024ehr} with static question-answer pairs $(x, y^*)$ or question-SQL pairs $(x, q^*)$, EHR-Complex derives the desired answer via multi-round interaction between the agent and the environment, \textit{i.e.}, EHR-Complex facilitates the scalable generation and sampling of diverse SQL trajectories $\{\tau^{(0)}, \dots, \tau^{(k)}\}$ with corresponding outputs $\{y^{(0)}, \dots, y^{(k)}\}$ via repeated independent agent rollouts.

\subsection{Benchmark Statistics}

\paragraph{Intent and Scope Taxonomy.}
EHR-Complex focuses on verifiable structured EHR tasks grounded in clinically motivated information needs and executable database evidence. Building on query types established in prior benchmarks~\citep{wang2020text,lee_ehrsql_2023,ryu2024ehr,jiang2025medagentbench}, we extend the scope to the full MIMIC-IV substrate across six clinical intents and two query scopes:
\begin{itemize}[leftmargin=*,nosep]
    \item \textbf{Intent:} Demographics \& Tracking (Demographics), Vital Signs \& Observations (Vitals), Medications \& Orders (Medications), Cost \& Billing (Cost), Labs \& Microbiology (Labs), and Diagnoses \& Procedures (Diagnoses).
    \item \textbf{Scope:} Patient-level (Pat.) and population-level (Pop.), corresponding to single-patient trajectory reasoning and cohort-level aggregation across patients, respectively.
\end{itemize}

Table~\ref{tab:data_stats} summarizes the EHR-Complex test-set distribution across these intents and query scopes. The full benchmark contains 48{,}092 training tasks and 3{,}915 test tasks. All reported evaluations are conducted on the test set.

\paragraph{SQL Task Complexity.}
We quantify SQL task complexity using the average number of SQL structural components in queries. Component definitions and details are provided in Appendix~\ref{sec:appendix_sql_complexity}. As shown in Table~\ref{tab:comparison} and Figure~\ref{fig:sql_complexity_dimensions}, EHR-Complex averages 31.93 components per query, reflecting the real-world SQL task complexity required by longitudinal multi-table aggregation and compositional reasoning.

\subsection{Data Curation Pipeline}
We formulate task construction as a grounded program synthesis problem over longitudinal EHR data. Let the underlying EHR database be denoted by $\mathcal{D}$. Our goal is to construct a synthesis mapping $f:\mathcal{D}\rightarrow (\mathcal{Q}, \mathcal{A}_{\mathrm{SQL}})$ that transforms raw patient records into a set of executable clinical queries $\mathcal{Q}$ and corresponding verifiable answers $\mathcal{A}_{\mathrm{SQL}}$.

\begin{table}[t]
\centering
\caption{Task Distribution of EHR-Complex Test Set.}
\label{tab:data_stats}
\small
\renewcommand{\arraystretch}{1.15}
\setlength{\tabcolsep}{10pt}
\begin{tabularx}{0.9\linewidth}{@{}>{\raggedright\arraybackslash}Xrrr@{}}
\toprule
\multirow{2}{*}{\raisebox{-0.35ex}{\textbf{Clinical intent}}}
& \multicolumn{2}{c}{\textbf{Query scope}}
& \multirow{2}{*}{\raisebox{-0.35ex}{\textbf{Total}}} \\
\cmidrule(lr){2-3}
& \textbf{Pop.} & \textbf{Pat.} & \\
\midrule
Labs         & 835 & 670 & 1{,}505 \\
Demographics & 491 & 460 & 951 \\
Diagnoses    & 410 & 240 & 650 \\
Medications  & 324 & 260 & 584 \\
Cost         & 40  & 90  & 130 \\
Vitals       & 55  & 40  & 95 \\
\midrule
\textbf{Total} & \textbf{2{,}155} & \textbf{1{,}760} & \textbf{3{,}915} \\
\bottomrule
\end{tabularx}
\end{table}

\paragraph{Step-1: Patient Event Graph.} 
To provide a unified substrate for compositional reasoning over patient trajectories, we induce a patient event graph from \textsc{MIMIC-IV}, serving as an intermediate structured representation for task generation. Formally, for each patient $p$, we construct a heterogeneous temporal graph $G_p=(V_p,E_p,\phi_V,\phi_E)$, where $V_p$ denotes clinically meaningful event nodes, $E_p\subseteq V_p\times V_p$ denotes typed relations between events, and $\phi_V,\phi_E$ are node-type and edge-type mappings, respectively. In this graph, heterogeneous records from admissions, ICU stays, laboratory events, diagnoses, prescriptions, and charted observations are normalized into event nodes $v=(c,t,a)$, where $c$ denotes the clinical concept, $t$ the timestamp, and $a$ the associated attributes. Temporal and semantic dependencies are then represented as edges $e=(v_i,r,v_j)$, where $r\in\mathcal{R}$ and $\mathcal{R}$ is a predefined set of relation types such as \texttt{precedes}, \texttt{indicates}, or \texttt{treated\_by}.

\paragraph{Step-2: Clinical Evidence Path.}
This step focuses on sampling concrete clinical evidence paths from the patient event graph, referring to a temporally ordered chain of clinically related events that supports a candidate query. An evidence path is defined as an ordered sequence
\[
P=(v_1,v_2,\dots,v_k), \quad (v_i,v_{i+1})\in E_p,
\]
where $t(v_1)\leq t(v_2)\leq \cdots \leq t(v_k)$ and the path satisfies both temporal consistency and clinical plausibility constraints. This path constructs a reasoning chain that adheres to clinical logic, encompassing diverse clinical events within the patient event graph, such as ICU stays, laboratory events, and charted observations. Hence, the system can not only track the timeline but also flexibly query clinical details at any moment. This formulation induces: (1) groundedness, every task is tied to an explicit chain of evidence in the source EHR, and (2) compositionality, complex tasks are generated by composing reusable reasoning operators over event paths instead of enumerating surface-form question patterns.

\paragraph{Step-3: Evidence-to-SQL Template.}
The clinical evidence paths are compiled into executable SQL templates using strong LLMs via a mapping $g: P \xmapsto{\mathrm{LLM}} s$, where $P$ is an evidence path and $s\in\mathcal{S}$ is a SQL program that operationalizes the intended reasoning process over $\mathcal{D}$. 
A patient-level query typically follows an evidence path for a single patient, whereas a population-level query is derived by aligning the same evidence-path pattern across multiple patients and aggregating the resulting evidence.
The synthesized query is then checked by database execution $\operatorname{Exec}(s,\mathcal{D})$. Specifically, we keep queries that execute successfully, return non-empty results, and admit deterministic answer extraction from the underlying records. This compilation step grounds high-level reasoning operators in database-level retrieval logic, while the subsequent execution check ensures that only well-formed tasks are retained. 

\paragraph{Template-based Scaling Up}
Data curation pipeline produces 796 structurally unique templates that are all validated by execution. We expand each template along two dimensions. First, we instantiate it in multiple concrete database contexts by resampling eligible patients and task-specific parameters such as time windows, diagnosis codes, laboratory items, medications, and target admissions, while keeping the underlying reasoning pattern fixed. Second, for each executable instance, we generate multiple independent natural language rewritings. This combination of structural reuse, patient-grounded variation, and linguistic diversity yields approximately 52K tasks, and every instance remains paired with an executable query and a verified exact answer.

\paragraph{Human Validation.} 
We conduct a two-stage human validation study on the test set to assess the semantic validity of EHR-Complex tasks.
First, we run four strong commercial model configurations on the full test set: GPT-5.4~\citep{gpt-5-4} with high and low reasoning effort, Gemini~3.1~Pro~\citep{gemini-3-1-pro}, and Claude~Sonnet~4.6~\citep{claude-sonnet-4-6}.
Each model independently interprets the natural-language query, interacts with the environment, and submits a final answer.
Among the 3{,}915 test tasks, 954 are not solved by any of these configurations, forming a challenging candidate set for human review.

Expert annotators then review a stratified sample of 500 tasks from this candidate set, focusing on \emph{clinical plausibility} and \emph{semantic alignment} rather than syntax alone.
Specifically, they assess whether the natural-language question faithfully reflects the clinical intent encoded by the SQL and whether the verified answer is supported by the retrieved database evidence.
The audit finds that 95.0\% of these challenging sampled tasks are valid after adjudication, providing evidence for the semantic reliability of the benchmark.
More details are reported in Appendix~\ref{sec:appendix_human_validation}.

\begin{table*}[t]
\centering
\setlength{\tabcolsep}{8pt} 
\renewcommand{\arraystretch}{1.58}

\caption{
Evaluation Results on the EHR-Complex Test Set.
}
\label{tab:main_results}

\resizebox{0.8\textwidth}{!}{%
\begin{tabular}{l *{12}{c} c}
\toprule
\multirow{2}{*}{\textbf{Model}}
& \multicolumn{2}{c}{\textbf{Demographics}}
& \multicolumn{2}{c}{\textbf{Vitals}}
& \multicolumn{2}{c}{\textbf{Medications}}
& \multicolumn{2}{c}{\textbf{Cost}}
& \multicolumn{2}{c}{\textbf{Labs}}
& \multicolumn{2}{c}{\textbf{Diagnoses}}
& \multirow{2}{*}{\textbf{Avg.}} \\
\cmidrule(lr){2-3}
\cmidrule(lr){4-5}
\cmidrule(lr){6-7}
\cmidrule(lr){8-9}
\cmidrule(lr){10-11}
\cmidrule(lr){12-13}
& \textbf{Pat.} & \textbf{Pop.}
& \textbf{Pat.} & \textbf{Pop.}
& \textbf{Pat.} & \textbf{Pop.}
& \textbf{Pat.} & \textbf{Pop.}
& \textbf{Pat.} & \textbf{Pop.}
& \textbf{Pat.} & \textbf{Pop.}
& \\
\midrule

\tabgroup{Proprietary API Models}
\model{GPT-4o} & \sv{51.5}{7.4} & \sv{10.4}{8.6} & \sv{45.0}{7.7} & \sv{25.5}{9.4} & \sv{32.7}{7.7} & \sv{7.1}{8.6} & \sv{53.3}{7.4} & \sv{32.5}{7.2} & \sv{45.5}{6.9} & \sv{19.0}{9.0} & \sv{38.3}{7.3} & \sv{7.6}{8.8} & \sv{30.7}{8.1} \\
\model{GPT-4.1 mini} & \sv{74.9}{9.1} & \sv{27.0}{8.6} & \sv{60.0}{10.7} & \sv{21.8}{7.1} & \sv{66.9}{9.1} & \sv{20.7}{8.7} & \sv{87.8}{8.5} & \sv{60.0}{6.8} & \sv{65.4}{8.7} & \sv{29.9}{8.8} & \sv{59.2}{8.7} & \sv{16.6}{9.0} & \sv{49.2}{8.8} \\
\model{GPT-4.1} & \sv{76.3}{9.2} & \sv{21.2}{9.9} & \sv{55.0}{10.7} & \sv{16.4}{8.4} & \sv{58.5}{8.9} & \sv{17.0}{10.3} & \sv{88.8}{8.1} & \sv{50.0}{8.6} & \sv{71.9}{7.6} & \sv{25.6}{9.9} & \sv{63.3}{8.2} & \sv{17.8}{10.6} & \sv{46.8}{9.3} \\
\model{Gemini 2.5 Pro} & \sv{43.5}{2.4} & \sv{18.5}{2.2} & \sv{17.5}{2.9} & \sv{20.0}{2.5} & \sv{42.7}{2.4} & \sv{14.9}{2.3} & \sv{58.9}{2.3} & \sv{37.5}{2.8} & \sv{38.4}{2.2} & \sv{27.2}{2.4} & \sv{34.2}{2.4} & \sv{14.9}{2.4} & \sv{30.7}{2.3} \\

\midrule
\tabgroup{Open-Weight Models $\ge$100B}
\model{Qwen3-235B} & \sv{74.8}{8.4} & \sv{27.1}{9.2} & \sv{65.0}{10.3} & \sv{25.5}{10.2} & \sv{71.8}{9.0} & \sv{24.5}{9.7} & \sv{91.1}{7.7} & \sv{55.0}{9.7} & \sv{79.7}{7.9} & \sv{36.1}{9.7} & \sv{70.0}{7.4} & \sv{20.5}{9.6} & \sv{53.4}{9.0} \\
\model{Qwen3.5-397B} & \sv{88.2}{9.8} & \sv{38.8}{13.1} & \sv{72.5}{12.7} & \sv{29.6}{12.8} & \sv{79.5}{11.7} & \sv{37.9}{13.4} & \sv{91.1}{8.2} & \sv{62.5}{11.7} & \sv{91.2}{8.6} & \sv{44.7}{12.9} & \sv{78.3}{9.3} & \sv{31.9}{13.0} & \sv{62.2}{11.4} \\
\model{DeepSeek-V3.1} & \sv{80.7}{6.5} & \sv{33.1}{8.1} & \sv{70.0}{8.3} & \sv{27.3}{8.3} & \sv{65.0}{7.5} & \sv{23.8}{8.5} & \sv{88.9}{6.2} & \sv{67.5}{7.6} & \sv{82.2}{5.9} & \sv{39.4}{8.2} & \sv{72.1}{6.0} & \sv{21.2}{8.1} & \sv{55.9}{7.4} \\
\model{DeepSeek-V3.2-Exp} & \sv{83.0}{6.7} & \sv{35.8}{12.0} & \sv{75.0}{9.6} & \sv{27.3}{11.9} & \sv{70.0}{9.7} & \sv{30.1}{13.0} & \sv{87.8}{5.3} & \sv{64.9}{9.6} & \sv{88.5}{6.4} & \sv{44.4}{11.1} & \sv{77.1}{6.8} & \sv{26.4}{11.3} & \sv{59.2}{9.5} \\
\model{Kimi-K2.5} & \sv{89.5}{10.3} & \sv{34.2}{11.8} & \sv{79.5}{14.5} & \sv{27.3}{13.2} & \sv{73.5}{12.0} & \sv{34.8}{12.4} & \sv{92.2}{8.7} & \sv{72.5}{10.4} & \sv{88.5}{8.6} & \sv{42.4}{11.8} & \sv{84.5}{9.3} & \sv{29.2}{11.5} & \sv{62.3}{10.9} \\

\midrule
\tabgroup{Open-Weight Models $<$100B}
\model{Qwen3-4B} & \sv{24.0}{9.5} & \sv{7.3}{14.9} & \sv{27.8}{11.8} & \sv{10.9}{14.7} & \sv{18.4}{8.5} & \sv{1.1}{15.3} & \sv{47.7}{9.7} & \sv{11.8}{11.6} & \sv{17.4}{9.5} & \sv{6.2}{13.1} & \sv{16.4}{8.0} & \sv{3.1}{15.3} & \sv{16.0}{11.9} \\
\model{Qwen3-14B} & \sv{44.1}{6.5} & \sv{14.5}{6.4} & \sv{20.5}{7.3} & \sv{18.2}{4.6} & \sv{43.0}{6.5} & \sv{15.4}{6.2} & \sv{62.9}{5.6} & \sv{33.3}{6.0} & \sv{40.3}{6.5} & \sv{19.2}{6.6} & \sv{38.5}{5.9} & \sv{9.7}{6.2} & \sv{30.0}{6.4} \\
\rowcolor{Emerald!8} \model{Qwen3-14B-SFT} & \sv{68.5}{28.5} & \sv{21.6}{34.0} & \sv{42.5}{31.7} & \sv{14.5}{33.0} & \sv{60.4}{28.9} & \sv{19.4}{37.3} & \sv{84.4}{23.0} & \sv{51.3}{28.3} & \sv{74.2}{24.4} & \sv{21.7}{37.0} & \sv{62.5}{30.2} & \sv{17.6}{36.1} & \sv{44.9}{31.9} \\
\model{Qwen3-32B} & \sv{51.0}{5.2} & \sv{20.3}{5.2} & \sv{30.0}{5.8} & \sv{23.6}{4.8} & \sv{46.9}{5.8} & \sv{19.8}{5.0} & \sv{73.3}{4.3} & \sv{35.0}{4.6} & \sv{49.6}{5.2} & \sv{21.0}{5.3} & \sv{42.3}{4.8} & \sv{15.0}{5.5} & \sv{35.7}{5.2} \\
\rowcolor{Emerald!8} \model{Qwen3-32B-SFT} & \sv{83.0}{12.0} & \sv{35.6}{17.2} & \sv{55.0}{20.2} & \sv{29.1}{19.8} & \sv{71.5}{14.0} & \sv{30.0}{18.0} & \sv{85.6}{10.3} & \sv{55.0}{13.1} & \sv{80.3}{11.0} & \sv{34.3}{18.1} & \sv{69.2}{12.5} & \sv{26.0}{19.3} & \sv{54.6}{15.4} \\

\bottomrule
\end{tabular}%
}

\vspace{0.5ex}
\begin{minipage}{0.8\textwidth}
\footnotesize
\colorbox{Emerald!8}{Green:} Models are fine-tuned on trajectories distilled from EHR-Complex train set (Section~\ref{sec:sft_validation}).
\end{minipage}
\end{table*}

\section{Experiments}

\subsection{Experimental Setup}

We evaluate 12 frontier LLMs on the EHR-Complex test set of 3{,}915 tasks. The closed-source APIs are GPT-4o-2024-11-20 (GPT-4o)~\citep{hurst2024gpt}, GPT-4.1-mini-2025-04-14 (GPT-4.1 mini), GPT-4.1-2025-04-14 (GPT-4.1)~\citep{gpt-4-1}, and Gemini-2.5-Pro (Gemini~2.5~Pro)~\citep{gemini-2-5-pro}. 
The open-weight models ($\ge$100B) are Qwen3-235B-A22B-Instruct-2507 (Qwen3-235B)~\citep{qwen3}, Qwen3.5-397B-A17B (Qwen3.5-397B)~\citep{qwen3-5}, DeepSeek-V3.1~\citep{deepseek-v3-1}, DeepSeek-V3.2-Exp~\citep{deepseek-v3-2-exp}, and Kimi-K2.5~\citep{kimi-k2-5}. 
The efficient-scale open-weight models ($<$100B) are Qwen3-4B, Qwen3-14B, and Qwen3-32B~\citep{qwen3}. All runs use a temperature of 0 and a maximum budget of 50 interaction turns, with each turn allowing one attempt to execute SQL/Python against the database and receive feedback. Task success is measured by an exact match against the verified ground truth answer. More implementation details are reported in Appendix~\ref{sec:appendix_implementation_details}.

\subsection{Main Results}

\paragraph{Overall Performance.}
Table~\ref{tab:main_results} reports evaluation success rates on EHR-Complex. Among 12 models evaluated without any fine-tuning, Kimi-K2.5 and Qwen3.5-397B achieve the highest macro-averaged scores, both around 0.62, leaving substantial room for improvement. Among proprietary models, GPT-4.1 mini and GPT-4.1 reach 0.49 and 0.47, respectively, while GPT-4o and Gemini~2.5~Pro both remain at 0.31. 
Performance also drops sharply at a smaller scale, with Qwen3-4B achieving only 0.16. These results suggest that EHR-Complex remains a challenging execution-based benchmark for current clinical agents.

\begin{figure*}[t]
  \centering
  \includegraphics[width=0.99\linewidth]{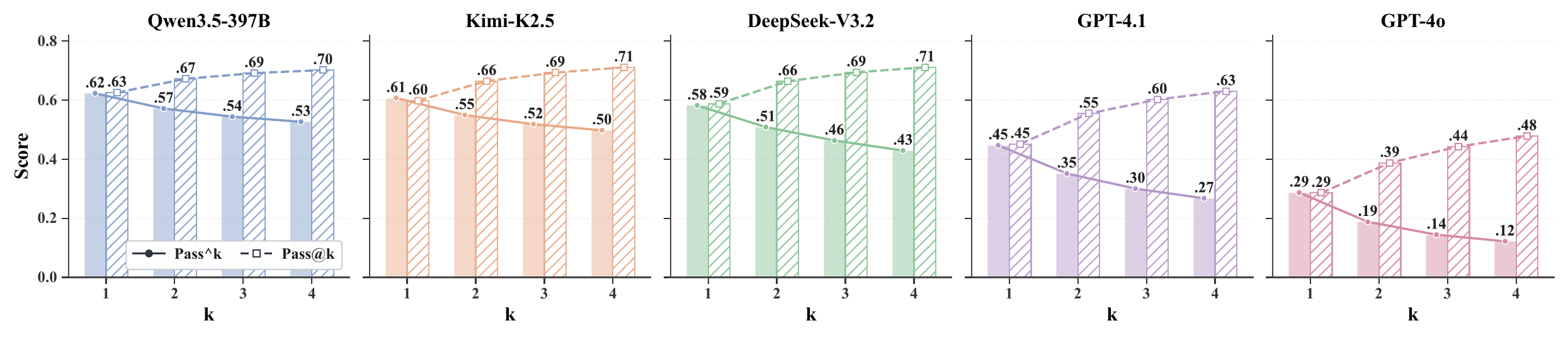}
\caption{\Passk{} and Pass@k at k = $1, 2, 3, 4$.}
  \label{fig:passk_all_models}
\end{figure*}

\begin{figure*}[t]
  \centering

  \begin{minipage}{0.99\textwidth}
    \centering
    \includegraphics[width=\linewidth]{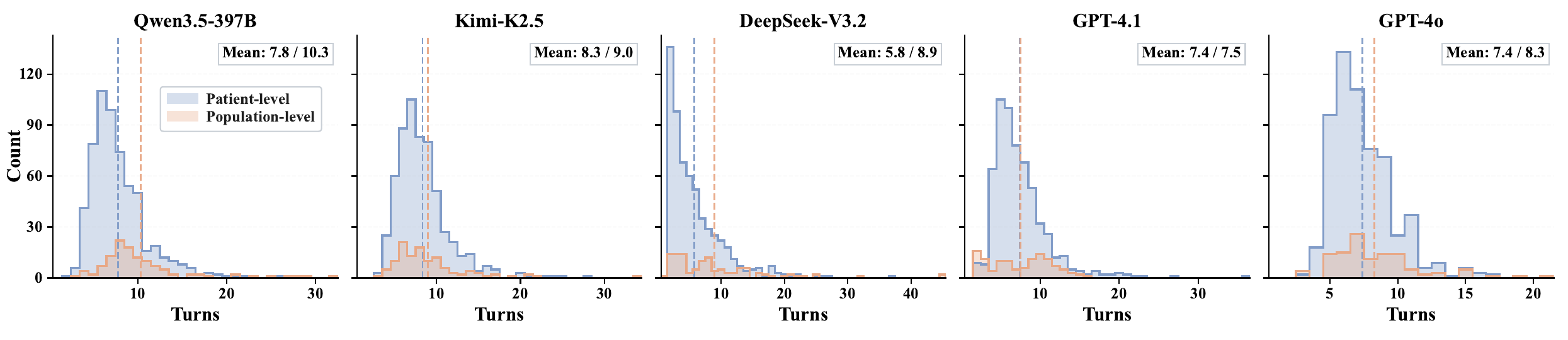}
    \par\vspace{0.2ex}
    {\small\textbf{(a)}~Turn Count Distribution across Successful Tasks.}
  \end{minipage}

  \vspace{1.0ex}

  \begin{minipage}{0.99\textwidth}
    \centering
    \includegraphics[width=\linewidth]{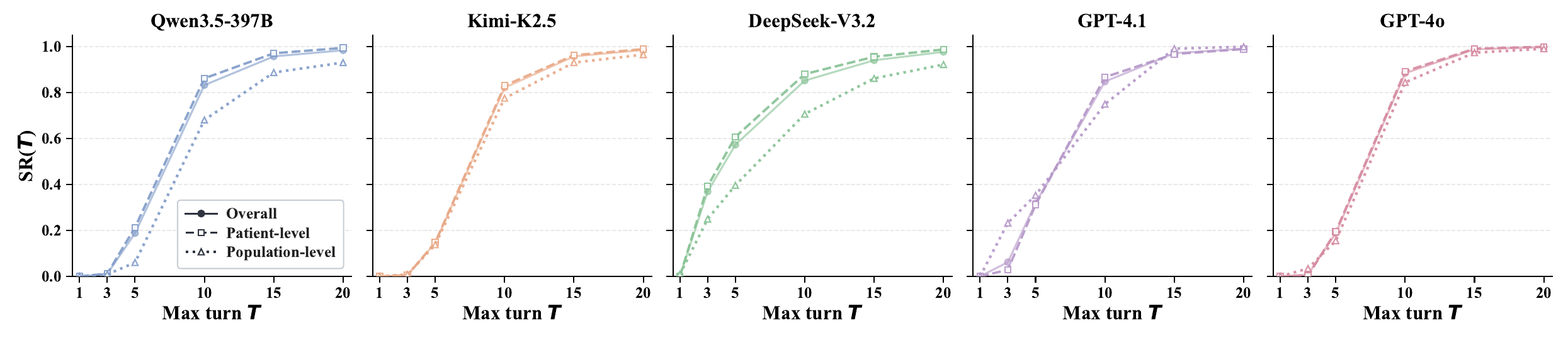}
    \par\vspace{0.2ex}
    {\small\textbf{(b)}~Success Rate (SR) When Limiting Max Turn $T$.}
  \end{minipage}

  \vspace{-0.2ex}
  \caption{Analysis on 712 Successful Tasks on EHR-Complex Test Set. These 712 trajectories are obtained by the intersection of the successful trajectories from five models, with the maximum turn budget set to 50.}
  \label{fig:interaction_turn_analysis}
\end{figure*}

\paragraph{Performance Gap between Patient- and Population-Level.}
\label{pag:pop_harder}
Averaging over clinical intents, Kimi-K2.5 drops from 0.85 to 0.40, Qwen3.5-397B from 0.83 to 0.41, and GPT-4.1 from 0.69 to 0.25 when moving from patient-level to population-level queries.
The gap is particularly pronounced for Diagnoses, where temporally grounded constraints are more common: Kimi-K2.5 drops from 0.85 to 0.29, and GPT-4.1 from 0.63 to 0.18.
This suggests that population-level queries introduce failures beyond single-patient reasoning: small errors in per-patient alignment, code grounding, or filtering can silently change the cohort before aggregation. Appendix~\ref{sec:appendix_cohort_failures} gives representative examples.

\paragraph{Cross Intent Analysis.}
At the patient level, Cost is among the highest-performing intents for strong models, with Kimi-K2.5 and GPT-4.1 reaching 0.92 and 0.89, respectively. Many Cost questions involve identifying the relevant admission or follow-up stay and retrieving admission-level attributes such as \texttt{insurance}.
Demographics and Labs also show relatively high success rates, but they require broader schema navigation: models must align admissions over time before accessing transfer, laboratory, or microbiology records. Vitals and Medications show lower success rates, as they involve denser event streams, temporal extrema, ordered measurements or medications, dosage constraints, and caregiver- or order-level aggregation.
At the population level, Cost remains comparatively strong, with Kimi-K2.5 at 0.73 and GPT-4.1 at 0.50. By contrast, Vitals, Diagnoses, and Medications are among the lowest-performing intents for Kimi-K2.5, at 0.27, 0.29, and 0.35, respectively.

\subsection{Additional Analysis}

\paragraph{\Passk{} and Pass@k}
Beyond the full-test evaluation in Table~\ref{tab:main_results}, we study whether model success is stable across repeated independent rollouts.
Because repeated execution is costly, we compute \Passk{} consistency and Pass@k on a stratified 20\% subsample of the EHR-Complex test set, comprising 783 tasks across nine representative models.
\Passk{} is evaluated at temperature 0.0 and requires all k rollouts to be correct, whereas Pass@k is evaluated at temperature 0.6 and requires at least one correct rollout, for k $=1,2,3,4$.

Figure~\ref{fig:passk_all_models} shows that multiple rollouts improve best-case success under Pass@k while exposing instability under \Passk{}.
Pass@k increases from 0.63 to 0.70 for Qwen3.5-397B and from 0.60 to 0.71 for Kimi-K2.5, indicating that some tasks are solved only when multiple trajectories are sampled.
By contrast, \Passk{} decreases with k: strong models drop from roughly 0.62 at k$=1$ to below 0.53 at k$=4$.
These results suggest that execution-based EHR reasoning remains sensitive to rollout-level variation, where small changes in temporal alignment, cohort definition, or aggregation logic can yield plausible but incorrect final answers.
Full model-, scope-, and intent-level results are reported in Appendix~\ref{sec:appendix_passk}.

\paragraph{Interaction Turn Analysis}
\label{sec:interaction_budget}

We analyze the 712 EHR-Complex test tasks solved correctly by all five representative models.
Because all five models solve these tasks under the 50-turn limit, this analysis isolates interaction efficiency rather than final solvability.

Figure~\ref{fig:interaction_turn_analysis}(a) shows that population-level tasks generally require more turns.
For Qwen3.5-397B, the mean turn count increases from 7.8 on patient-level tasks to 10.3 on population-level tasks. DeepSeek-V3.2-Exp uses the fewest turns on patient-level tasks, but this advantage narrows on population-level aggregation.
For each maximum-turn limit $T \in \{1,3,5,10,15,20\}$, we define $\mathrm{SR}(T)$ as the fraction of the 712 tasks completed within $T$ turns.
Figure~\ref{fig:interaction_turn_analysis}(b) shows that small turn limits are insufficient, while SR increases sharply by $T=10$ and largely plateaus by $T=15$.
Full results are reported in Appendix~\ref{sec:appendix_budget}.

\paragraph{Failure Case Analysis}
\label{sec:comparative_error}

\begin{figure}[t]
  \centering
  \includegraphics[width=\linewidth]{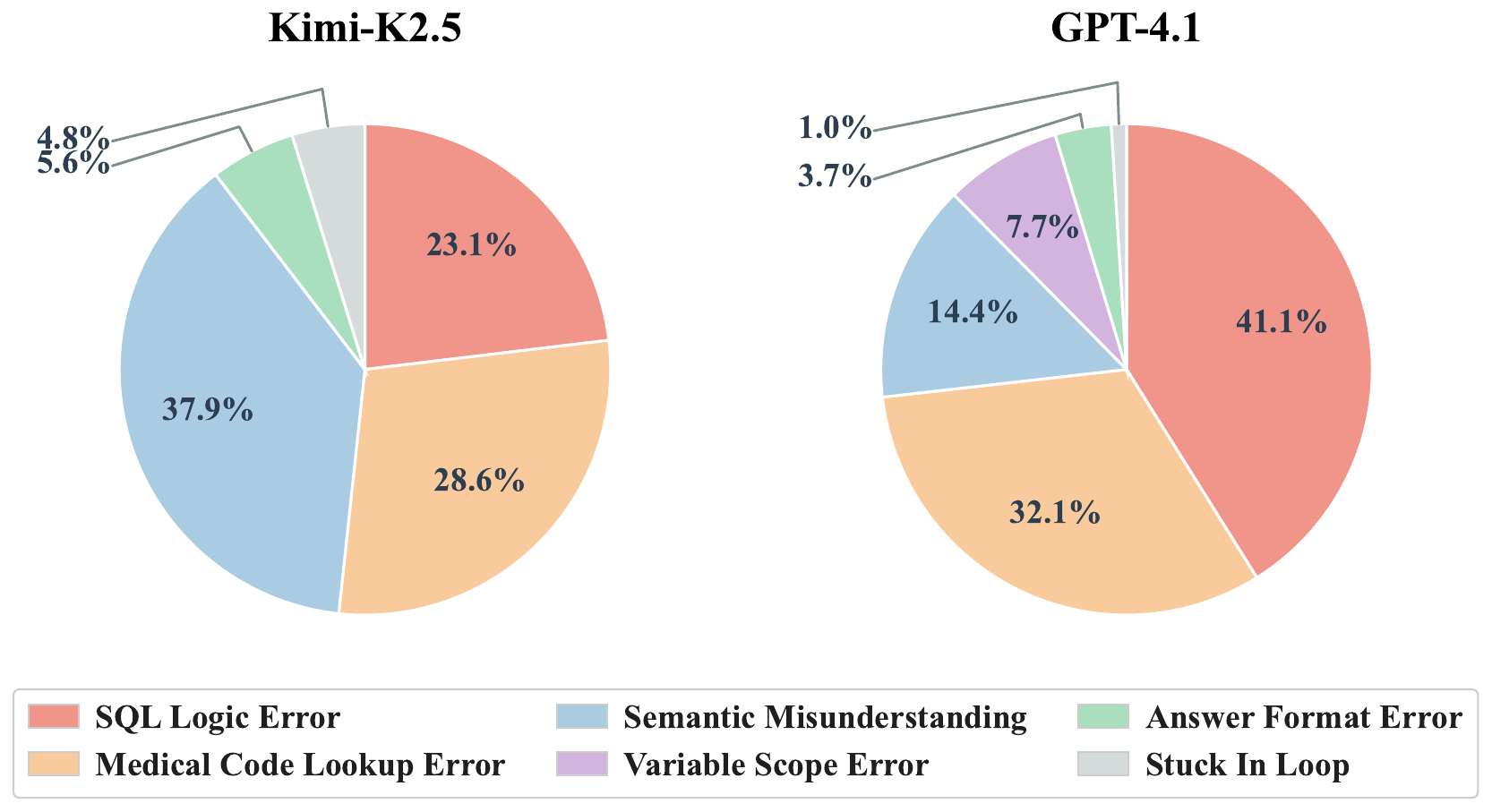}
  \caption{Distribution of Failure Categories.}
  \label{fig:error_distribution}
\end{figure}

To understand failure patterns on EHR-Complex, we analyze 3{,}814 failed trajectories from two representative models: Kimi-K2.5 and GPT-4.1.
We first manually inspect more than 50 failures per model to derive a compact taxonomy, and then use Gemini~2.5~Pro to label the remaining failures at scale under the same rubric.
The taxonomy contains six categories: \emph{Semantic Misunderstanding}, 
\emph{Medical-Code Lookup Error},  
\emph{SQL Logic Error}, 
\emph{Variable Scope Error}, 
\emph{Answer Format Error} 
and \emph{Stuck-in-Loop}. 
Figure~\ref{fig:error_distribution} shows that Kimi-K2.5 is dominated by \emph{Semantic Misunderstanding} (37.9\%) and \emph{Medical-Code Lookup Error} (28.6\%), indicating that many failures originate in clinical interpretation or grounding.
GPT-4.1, by contrast, is dominated by \emph{SQL Logic Error} (41.1\%) and \emph{Medical-Code Lookup Error} (32.1\%), suggesting that it often captures the broad task intent but loses correctness during query construction.
Further details are provided in Appendix~\ref{sec:appendix_error_analysis}.

\subsection{Trajectory Supervision Training}
\label{sec:sft_validation}


We evaluate whether successful EHR-Complex trajectories offer effective supervision for interactive reasoning. Distilling 6,000 Kimi-K2.5-generated trajectories from the training set (with no test overlap), we fine-tune Qwen3-14B and Qwen3-32B via full-parameter SFT for one epoch. As shown in Table~\ref{tab:main_results}, this yields consistent in-domain gains: Qwen3-14B improves from 0.30 to 0.45, and Qwen3-32B from 0.36 to 0.55. Notably, the fine-tuned 32B model surpasses GPT-4.1 and becomes competitive with the much larger Qwen3-235B. These results confirm that EHR-Complex trajectories provide learnable procedural supervision for schema navigation and temporal grounding, validating the environment as a scalable source for agent training. Full details are in Appendix~\ref{sec:appendix_sft}.

\section{Conclusion}

We introduced EHR-Complex, a large-scale benchmark for interactive clinical database reasoning over longitudinal EHRs. Built on the full MIMIC-IV substrate, EHR-Complex covers diverse patient- and population-level queries that require sandboxed SQL/Python execution, multi-table aggregation, temporal reasoning, and compositional SQL logic. Experiments with broad LLM agents reveal persistent limitations in robust EHR reasoning, especially for population-level reasoning, medical-code grounding, and clinically faithful SQL construction, highlighting the gap between current capabilities and dependable large-scale EHR analysis.

\section*{Limitations}

EHR-Complex is built on MIMIC-IV v3.1, a large-scale publicly available EHR database released for research under credentialed access. This inevitably bounds the benchmark to a single hospital system, including its EHR schema, ICU information system, coding conventions, patient population, and clinical practice patterns; therefore, results may not directly transfer to other hospitals, EHR vendors, or international clinical settings. At the same time, MIMIC-IV provides a reproducible and privacy-preserving substrate for clinical-agent research: it has undergone professional de-identification, identifier replacement, PHI removal, and patient-level date shifting. This enables us to release the full EHR-Complex train/test task definitions, SQL programs, evaluation environment, and construction code, while requiring users to access the underlying MIMIC-IV records through the standard credentialed-access procedure. EHR-Complex focuses on structured database reasoning and does not cover free-text notes, imaging data, real-time clinician interaction, or prospective clinical deployment.


\bibliography{custom}

\clearpage
\appendix

\section{Ethical Considerations}
\label{sec:appendix_ethics}

\paragraph{Data privacy and access.}
EHR-Complex is constructed from MIMIC-IV v3.1, a large-scale de-identified EHR database released for research under credentialed access. MIMIC-IV is derived from routine clinical records at Beth Israel Deaconess Medical Center and integrates hospital-wide EHR data with ICU-specific MetaVision records. The underlying data have undergone privacy-preserving processing, including HIPAA identifier removal, random identifier replacement, structured-data filtering, free-text PHI removal, and date shifting. We use MIMIC-IV only for research purposes and do not redistribute raw patient records, identifiers, or restricted database files. Any released EHR-Complex artifacts, such as task definitions, SQL programs, evaluation code, and construction scripts, will be shared only in forms compatible with the original MIMIC-IV access conditions.

\paragraph{Access requirements and user obligations.}
Access to the underlying MIMIC-IV records requires users to complete the official PhysioNet credentialing process and the required CITI Data or Specimens Only Research training, and to sign the PhysioNet Credentialed Health Data Use Agreement. Each user must obtain individual access rights, and access must not be shared with anyone else. Users must access the underlying EHR records only through the official credentialed-access procedure, use them solely for lawful scientific research, maintain appropriate data security, and must not attempt to re-identify patients, institutions, or any other protected entities.

\paragraph{Artifact use and licensing.}
We use datasets, models, and software artifacts under their respective licenses and access terms, and cite their creators where applicable. The derived benchmark is intended to support reproducible research on medical agents and clinical database reasoning while respecting the restrictions associated with the underlying EHR data. In particular, releasing EHR-Complex task definitions and evaluation code does not grant access to MIMIC-IV itself. Users remain responsible for satisfying the original PhysioNet credentialing, training, and data-use requirements before executing benchmark tasks against restricted EHR records.

\paragraph{Intended use and potential risks.}
EHR-Complex is intended as a research benchmark for evaluating and training agents on structured clinical database reasoning, not as a clinical decision-support system. Because it is derived from a single-institution de-identified dataset, benchmark results should be interpreted as evidence of database reasoning ability rather than as direct evidence of clinical readiness. Access to the underlying MIMIC-IV records remains governed by the official PhysioNet credentialing, required training, and data-use agreement process, so released task artifacts do not by themselves provide access to restricted EHR data. We emphasize responsible research use, compliance with MIMIC-IV access conditions, and independent validation before any clinical application.

\paragraph{Use of AI assistants.}
AI assistants were used to support language polishing. All scientific claims, experimental results, analyses, citations, and final paper content were reviewed and verified by the authors.

\section{Implementation Details}
\label{sec:appendix_implementation_details}

All evaluated systems use the same EHR-Complex agent interface.
The agent receives the database schema description, the clinical question, and access to the executable tool, \texttt{execute\_code}. Each tool call executes isolated SQL query or Python code. No variables are shared across tool calls. The agent must submit intermediate tool calls as valid JSON and return the final prediction inside a \texttt{<Final\_answer>} tag.
The SQL and verified answer are never exposed to the model during inference and are used only after submission for exact-match validation.

The implementation supports OpenAI, Azure, and local vLLM-style clients. All reported EHR-Complex runs use an OpenAI-compatible chat-completion API format.
This format normalizes proprietary API calls and local or internally hosted open-weight endpoints under the same request schema.
Transient API errors are retried by the model wrapper, invalid JSON or parsing errors are returned as environment feedback, and the environment enforces the 50-turn interaction budget used in the main experiments.
Except for the Pass@k ablations, evaluation uses temperature 0.0.
For locally served Qwen3 models, the client passes \texttt{enable\_thinking} through both standard and SGLang-compatible request fields where supported.
Table~\ref{tab:model_invocation_details} lists the model identifiers and invocation settings used in our experiments. All open-weight models are served with SGLang.

\begin{table*}[htbp]
\centering
\small
\renewcommand{\arraystretch}{1.10}
\setlength{\tabcolsep}{4pt}
\begin{tabularx}{\textwidth}{@{}p{0.30\linewidth}>{\ttfamily\footnotesize\raggedright\arraybackslash}X p{0.18\linewidth} p{0.13\linewidth}@{}}
\toprule
\textbf{Model} & \textbf{Run identifier} & \textbf{Backend} & \textbf{Max output} \\
\midrule
\multicolumn{4}{@{}l}{\textit{Proprietary API models}} \\
GPT-4o & gpt-4o-2024-11-20 & API & 16{,}384 \\
GPT-4.1 mini & gpt-4.1-mini-2025-04-14 & API & 32{,}768 \\
GPT-4.1 & gpt-4.1-2025-04-14 & API & 32{,}768 \\
Gemini 2.5 Pro & gemini-2.5-pro & API & 65{,}536 \\
\midrule
\multicolumn{4}{@{}l}{\textit{Open-weight models $\ge$100B}} \\
Qwen3-235B & Qwen3-235B-A22B-Instruct-2507 & SGLang & 65{,}536 \\
Qwen3.5-397B & Qwen3.5-397B-A17B & SGLang & 65{,}536 \\
DeepSeek-V3.1 & DeepSeek-V3.1 & SGLang & 65{,}536 \\
DeepSeek-V3.2-Exp & DeepSeek-V3.2-Exp & SGLang & 65{,}536 \\
Kimi-K2.5 & Kimi-K2.5 & SGLang & 65{,}536 \\
\midrule
\multicolumn{4}{@{}l}{\textit{Open-weight models $<$100B}} \\
Qwen3-4B & Qwen3-4B & SGLang & 32{,}768$^{\dagger}$ \\
Qwen3-14B & Qwen3-14B & SGLang & 32{,}768$^{\dagger}$ \\
Qwen3-32B & Qwen3-32B & SGLang & 32{,}768$^{\dagger}$ \\
\bottomrule
\end{tabularx}
\vspace{0.3ex}
\begin{minipage}{0.98\textwidth}
\footnotesize
$^{\dagger}$ Thinking is enabled for locally served Qwen3 models.
\end{minipage}
\caption{Model invocation settings.}
\label{tab:model_invocation_details}
\end{table*}

\section{Template Materialization Examples}
\label{sec:appendix_template_materialization}

Table~\ref{tab:sql_template_structure} summarizes one representative SQL template before parameter materialization.
The template anchors an initial admission, retrieves later admissions, and aggregates the elapsed time after a subsequent laboratory event. Table~\ref{tab:param_expansion_examples} shows parameter-expanded instances from this template.
Each row changes the clinical concepts and database constants while preserving the same executable reasoning pattern. Table~\ref{tab:paraphrase_examples} illustrates natural-language rewritings for one executable instance.
All rewritings preserve the same temporal anchor, target measurement, and verified answer.

\section{SQL Structural Component Definitions}
\label{sec:appendix_sql_complexity}

SQL task complexity is quantified with an AST-based extraction procedure implemented using \texttt{sqlglot}. For each query, we parse the SQL program and extract the structural components listed in Table~\ref{tab:sql_component_definitions}, including nesting, aggregation, joins, conditions, grouping or ordering, and window functions.
These components are summed within each query to obtain a per-query structural total, and the reported complexity is the average of these totals over tasks.
We apply the same extraction procedure to all datasets in Table~\ref{tab:comparison} and Figure~\ref{fig:sql_complexity_dimensions}.

Table~\ref{tab:sql_complexity_finegrained} reports the fine-grained SQL structural component counts for EHR-Complex test tasks. TOTAL is computed before rounding.

\begin{table*}[t]
\centering
\small
\renewcommand{\arraystretch}{1.14}
\setlength{\tabcolsep}{4pt}
\begin{tabularx}{\textwidth}{@{}p{0.24\linewidth}X@{}}
\toprule
\textbf{Template stage} & \textbf{Executable logic} \\
\midrule
Initial admission candidates
& Join \texttt{admissions}, \texttt{diagnoses\_icd}, and \texttt{labevents}; retain admissions matching the initial diagnosis placeholder \texttt{[ICD\_CODE\_INITIAL]} and initial laboratory placeholder \texttt{[LAB\_ITEM\_ID\_INITIAL]}. \\
\addlinespace[0.35ex]
First qualifying admission
& Rank admissions by \texttt{admittime} within each patient and keep the earliest qualifying admission as the temporal anchor. \\
\addlinespace[0.35ex]
Subsequent admissions
& Retrieve later admissions for the same patient by requiring \texttt{admittime} to be later than the temporal anchor. \\
\addlinespace[0.35ex]
Subsequent laboratory filter
& Join later admissions with \texttt{labevents} and retain admissions containing the subsequent laboratory placeholder \texttt{[LAB\_ITEM\_ID\_SUBSEQUENT]}. \\
\addlinespace[0.35ex]
Population aggregation
& Compute the day difference between the later admission and the initial admission, then return the average interval as the verified answer. \\
\bottomrule
\end{tabularx}
\caption{Representative SQL template structure.}
\label{tab:sql_template_structure}
\end{table*}

\begin{table*}[!t]
\centering
\small
\renewcommand{\arraystretch}{1.16}
\setlength{\tabcolsep}{4pt}
\begin{tabularx}{\textwidth}{@{}p{0.31\linewidth}X p{0.13\linewidth}@{}}
\toprule
\textbf{Parameter expansion} & \textbf{Generated clinical question} & \textbf{Verified answer} \\
\midrule
PHPT (\texttt{25201}); Total Chol (\texttt{50907}); Alk Phos (\texttt{50863})
& Among patients whose first qualifying admission carried diagnosis code PHPT and included lab Total Chol, what is the population-level mean days until a subsequent hospitalization where Alk Phos was drawn?
& 439.40 days \\
\addlinespace[0.35ex]
HLD (\texttt{E785}); Baso (\texttt{51146}); TC/HDL ratio (\texttt{50903})
& For the whole patient cohort, calculate the average time elapsed in days from an initial admission with diagnosis HLD and lab Baso to a subsequent hospitalization featuring lab TC/HDL ratio.
& 609.53 days \\
\addlinespace[0.35ex]
Liver disease (\texttt{K7689}); G6PD (\texttt{52123}); Retic count (\texttt{51282})
& What is the average days separating an initial admission defined by Liver disease and G6PD from a later hospitalization where Retic count was tested, across all patients?
& 84.00 days \\
\addlinespace[0.35ex]
h/o tobacco use (\texttt{V1582}); Total/HDL Ratio (\texttt{50903}); Sed Rate (\texttt{51288})
& Looking at the entire patient panel, what is the average duration in days separating an index h/o tobacco use admission with a Total/HDL Ratio baseline and a later readmission involving a Sed Rate test?
& 457.27 days \\
\bottomrule
\end{tabularx}
\caption{Examples of parameter-expanded instances.}
\label{tab:param_expansion_examples}
\end{table*}

\begin{table*}[!t]
\centering
\small
\renewcommand{\arraystretch}{1.12}
\setlength{\tabcolsep}{4pt}
\begin{tabularx}{\textwidth}{@{}p{0.09\linewidth}X@{}}
\toprule
\textbf{Rewrite} & \textbf{Question} \\
\midrule
1 & What was the nadir hematocrit during the second admission for patient 10000032? \\
2 & Can we identify the lowest Hct recorded during patient 10000032's second hospitalization? \\
3 & I need the lowest hematocrit value from the second admission for patient 10000032. \\
4 & What was the trough Hct during the second hospitalization for patient 10000032? \\
5 & Please provide the minimum hematocrit reading from patient 10000032's second hospital encounter. \\
6 & What is the trough hematocrit value for patient 10000032 during admission number two? \\
\bottomrule
\end{tabularx}
\caption{Natural-language rewritings for one executable instance.}
\label{tab:paraphrase_examples}
\end{table*}

\begin{table*}[!t]
\centering
\small
\renewcommand{\arraystretch}{1.12}
\setlength{\tabcolsep}{4pt}
\begin{tabularx}{\linewidth}{@{}p{0.20\linewidth}p{0.31\linewidth}X@{}}
\toprule
\textbf{Component} & \textbf{Operationalization} & \textbf{Interpretation} \\
\midrule
NEST
& Number of \texttt{SELECT} clauses minus one.
& Captures nested queries and common table expressions, reflecting multi-step retrieval or intermediate logic. \\

AGG
& Number of aggregate function calls, including \texttt{COUNT}, \texttt{SUM}, \texttt{AVG}, \texttt{MIN}, and \texttt{MAX}.
& Measures statistical summarization over raw records, such as counting patients or averaging measurements. \\

JOIN
& Sum of explicit joins, implicit cross-table equality links, and subquery-based associations such as \texttt{IN (SELECT ...)}.
& Reflects how many database tables or event sources must be connected to answer the question. \\

COND
& Number of filtering and logical clauses, including \texttt{WHERE}, \texttt{HAVING}, \texttt{AND}, and \texttt{OR}.
& Captures the granularity of clinical constraints, such as time windows, diagnosis filters, item identifiers, or cohort restrictions. \\

GROUP/ORDER
& Number of \texttt{GROUP BY} and \texttt{ORDER BY} clauses.
& Reflects result-shaping operations such as ranking, top-k retrieval, or aggregation by patient, admission, unit, or category. \\

WINDOW
& Number of \texttt{OVER} clauses, covering window functions such as \texttt{ROW\_NUMBER}, \texttt{RANK}, \texttt{LEAD}, and \texttt{LAG}.
& Captures temporal or within-group reasoning, such as selecting the first admission, latest measurement, or ordered event. \\

TOTAL
& Sum of the six reported component counts for each query.
& Used as the per-query structural complexity value before averaging over tasks. \\
\bottomrule
\end{tabularx}
\caption{Definitions of SQL structural components used in the complexity analysis.}
\label{tab:sql_component_definitions}
\end{table*}

\begin{table*}[!th]
\centering
\scriptsize
\renewcommand{\arraystretch}{1.10}
\setlength{\tabcolsep}{3pt}
\resizebox{0.98\textwidth}{!}{%
\begin{tabular}{llrrrrrrrr}
\toprule
\textbf{Scope} & \textbf{Clinical intent} & \textbf{Tasks} & \textbf{NEST} & \textbf{AGG} & \textbf{JOIN} & \textbf{COND} & \textbf{GROUP/ORDER} & \textbf{WINDOW} & \textbf{TOTAL} \\
\midrule
All & Overall & 3{,}915 & 10.27 & 1.36 & 4.23 & 10.65 & 3.95 & 1.46 & 31.93 \\
\midrule
Pop. & Overall & 2{,}155 & 9.87 & 1.42 & 4.12 & 9.75 & 4.07 & 1.42 & 30.65 \\
Pop. & Labs & 835 & 10.28 & 1.58 & 3.97 & 9.58 & 4.36 & 1.47 & 31.24 \\
Pop. & Demographics & 491 & 9.56 & 1.30 & 4.12 & 9.77 & 4.01 & 1.43 & 30.18 \\
Pop. & Diagnoses & 410 & 9.45 & 1.24 & 4.33 & 9.80 & 3.67 & 1.33 & 29.82 \\
Pop. & Medications & 324 & 10.01 & 1.37 & 4.23 & 10.26 & 3.95 & 1.52 & 31.33 \\
Pop. & Cost & 40 & 8.50 & 1.63 & 4.25 & 8.25 & 3.38 & 0.38 & 26.38 \\
Pop. & Vitals & 55 & 9.64 & 1.55 & 4.27 & 10.00 & 4.18 & 1.36 & 31.00 \\
\midrule
Pat. & Overall & 1{,}760 & 10.76 & 1.29 & 4.36 & 11.76 & 3.82 & 1.51 & 33.49 \\
Pat. & Labs & 670 & 10.82 & 1.57 & 4.02 & 11.57 & 3.48 & 1.40 & 32.86 \\
Pat. & Demographics & 460 & 10.79 & 1.20 & 4.40 & 11.27 & 3.86 & 1.54 & 33.07 \\
Pat. & Diagnoses & 240 & 10.94 & 0.83 & 5.06 & 12.13 & 4.04 & 1.65 & 34.65 \\
Pat. & Medications & 260 & 10.98 & 1.15 & 4.60 & 12.79 & 4.15 & 1.63 & 35.31 \\
Pat. & Cost & 90 & 9.50 & 1.11 & 4.11 & 11.39 & 4.28 & 1.39 & 31.78 \\
Pat. & Vitals & 40 & 9.50 & 1.88 & 4.50 & 12.50 & 4.38 & 1.50 & 34.25 \\
\bottomrule
\end{tabular}
}
\caption{Fine-grained SQL structural component counts for EHR-Complex test tasks.}
\label{tab:sql_complexity_finegrained}
\end{table*}

\section{Human Validation of Task Quality}
\label{sec:appendix_human_validation}

Execution-based filtering verifies that a synthesized SQL program can be executed successfully and yields a deterministic answer.
However, successful execution alone does not guarantee that the natural-language question, SQL logic, and final answer are semantically aligned.
We therefore perform a two-stage task-quality validation procedure: a strong-model stress test over the full test set followed by an expert human audit on challenging cases.

\paragraph{Strong-model stress test.}
Before human review, we run four strong commercial model configurations from three model families: GPT-5.4~\citep{gpt-5-4} with high and low reasoning effort, Gemini~3.1~Pro~\citep{gemini-3-1-pro}, and Claude~Sonnet~4.6~\citep{claude-sonnet-4-6}.
For each test task, the model receives the same agent interface as evaluated models, including the natural-language question, schema information, and executable SQL/Python environment.
The reference SQL program and verified answer are not exposed to the model.
Each model independently interprets the question, interacts with the environment, and submits a final answer for exact-match evaluation.

Table~\ref{tab:strong_model_results} reports the strong-model results on the EHR-Complex test set.
This stage is used to identify challenging cases rather than to certify task validity.
Among the 3{,}915 test tasks, 954 are not solved by any of the four configurations.
We use these 954 unsolved tasks as the candidate pool for human review, because they concentrate the cases most likely to expose either genuine task difficulty or potential semantic issues.

\paragraph{Sampling and annotation.}
From the 954 unsolved candidate tasks, we sample 500 task instances for human validation.
The sample is stratified across the six clinical intents and two query scopes, with allocation proportional to the number of retained instances in each intent--scope cell.
To avoid auditing only repetitive cases, we also preserve variation in SQL structure and task templates during sampling.
The audited tasks cover 390 structurally unique templates.

Each task is independently reviewed by two annotators with medical AI backgrounds, as well as experience in EHR data analysis and SQL-based database querying.
Annotators are shown the natural-language question, reference SQL query, execution result, extracted final answer, relevant schema information, and supporting database evidence used during task construction.
They focus on clinical plausibility and semantic alignment rather than SQL syntax alone.
Specifically, they assess whether the natural-language question faithfully reflects the clinical intent encoded by the SQL and whether the extracted ground-truth answer is supported by the retrieved evidence.
The human annotation criteria are summarized in Figure~\ref{fig:human_annotation_criteria}.

Annotators assign a binary validity label: 1 indicates that no substantive issue is found, and 0 indicates a concern requiring further inspection.
They are blinded to strong-model predictions and downstream failure analyses.
Disagreements are resolved through joint adjudication after inspecting the question, SQL, execution result, and supporting evidence.

\paragraph{Results.}
Table~\ref{tab:human_validation} summarizes the human validation results.
After adjudication, 475 of the 500 audited unsolved tasks are judged valid, corresponding to a validity rate of 95.0\%.
The most common concerns involve borderline temporal wording, cohort-unit interpretation, medical-code granularity, or answer-format mismatches.
Because the audited set is sampled from tasks unsolved by all four strong configurations, this result provides a conservative check of benchmark validity on difficult cases.
The high validity rate suggests that most strong-model failures in this subset reflect task difficulty rather than invalid task construction.

\paragraph{Inter-annotator agreement.}
Before adjudication, the two annotators agree on 465 of the 500 binary labels, yielding 93.0\% raw agreement.
Because both annotators mark approximately 95\% of tasks as valid, Cohen's $\kappa$ is reduced by the strong label-prevalence imbalance ($\kappa=0.25$).
The prevalence-adjusted bias-adjusted kappa (PABAK) is 0.86.
Disagreements are primarily caused by borderline temporal wording, alternative interpretations of cohort units, or medical-code granularity, and are resolved through adjudication.

\begin{table*}[t]
\centering
\setlength{\tabcolsep}{8pt}
\renewcommand{\arraystretch}{1.58}

\caption{Strong commercial model results on the EHR-Complex test set.}
\label{tab:strong_model_results}

\resizebox{0.8\textwidth}{!}{%
\begin{tabular}{l *{12}{c} c}
\toprule
\multirow{2}{*}{\textbf{Model}}
& \multicolumn{2}{c}{\textbf{Demographics}}
& \multicolumn{2}{c}{\textbf{Vitals}}
& \multicolumn{2}{c}{\textbf{Medications}}
& \multicolumn{2}{c}{\textbf{Cost}}
& \multicolumn{2}{c}{\textbf{Labs}}
& \multicolumn{2}{c}{\textbf{Diagnoses}}
& \multirow{2}{*}{\textbf{Avg.}} \\
\cmidrule(lr){2-3}
\cmidrule(lr){4-5}
\cmidrule(lr){6-7}
\cmidrule(lr){8-9}
\cmidrule(lr){10-11}
\cmidrule(lr){12-13}
& \textbf{Pat.} & \textbf{Pop.}
& \textbf{Pat.} & \textbf{Pop.}
& \textbf{Pat.} & \textbf{Pop.}
& \textbf{Pat.} & \textbf{Pop.}
& \textbf{Pat.} & \textbf{Pop.}
& \textbf{Pat.} & \textbf{Pop.}
& \\
\midrule

\model{GPT-5.4 (high)} & \sv{85.4}{2.9} & \sv{36.0}{4.7} & \sv{77.5}{4.3} & \sv{34.5}{3.9} & \sv{80.4}{3.7} & \sv{37.0}{4.8} & \sv{93.3}{2.4} & \sv{76.3}{4.2} & \sv{91.2}{3.1} & \sv{49.0}{4.8} & \sv{84.2}{2.9} & \sv{35.6}{4.3} & \sv{65.0}{3.9} \\
\model{GPT-5.4 (low)} & \sv{81.1}{2.5} & \sv{32.3}{3.0} & \sv{65.0}{4.0} & \sv{25.9}{2.9} & \sv{72.7}{3.1} & \sv{30.8}{3.1} & \sv{88.3}{2.3} & \sv{66.1}{2.9} & \sv{85.9}{2.7} & \sv{43.4}{3.1} & \sv{79.6}{2.6} & \sv{28.7}{3.2} & \sv{58.3}{2.9} \\
\model{Gemini 3.1 Pro} & \sv{86.7}{7.8} & \sv{32.3}{21.0} & \sv{67.5}{12.9} & \sv{38.2}{14.5} & \sv{83.8}{12.3} & \sv{37.0}{19.4} & \sv{92.2}{5.8} & \sv{62.5}{20.1} & \sv{92.5}{8.5} & \sv{43.8}{18.9} & \sv{82.5}{7.7} & \sv{33.6}{20.0} & \sv{62.7}{14.1} \\
\model{Claude Sonnet 4.6} & \sv{59.4}{4.5} & \sv{20.7}{3.7} & \sv{45.0}{8.4} & \sv{12.0}{3.7} & \sv{45.4}{6.1} & \sv{12.4}{3.6} & \sv{83.2}{4.5} & \sv{20.0}{4.3} & \sv{48.9}{4.9} & \sv{33.3}{3.1} & \sv{40.3}{4.5} & \sv{10.8}{3.4} & \sv{36.0}{4.6} \\

\bottomrule
\end{tabular}%
}

\vspace{0.5ex}
\end{table*}

\begin{figure*}
\begin{tcolorbox}[
  enhanced,
  colback=casebeige,
  colframe=casebeige!70!black,
  boxrule=0.4pt,
  arc=2pt,
  left=6pt,
  right=6pt,
  top=5pt,
  bottom=5pt,
  title={Human annotation criteria},
  fonttitle=\bfseries
]
A task is marked as \emph{valid} only if it satisfies all four checks below.
Minor wording issues that do not affect the intended clinical meaning, SQL logic, or final answer are recorded separately but are not counted as invalid.

\vspace{0.5ex}
\small
\renewcommand{\arraystretch}{1.08}
\begin{tabularx}{\linewidth}{@{}p{0.28\linewidth}X@{}}
\toprule
\textbf{Check} & \textbf{Requirement} \\
\midrule
Question clarity
& The natural-language question is understandable, clinically plausible, and answerable from the provided EHR database. \\
Semantic alignment
& The SQL faithfully implements the intent expressed in the natural-language question, including joins, filters, temporal constraints, aggregation units, and cohort definitions. \\
Medical grounding
& Clinical concepts, medical codes, laboratory items, medications, procedures, and diagnoses are grounded to appropriate database identifiers. \\
Answer support
& The SQL execution result and supporting evidence justify a unique final answer under the benchmark's answer-extraction rule. \\
\bottomrule
\end{tabularx}

\vspace{0.5ex}
Major issues include ambiguous clinical questions, incorrect medical-code grounding, wrong temporal constraints, incorrect aggregation units, mismatches between the SQL result and the final answer, or answers that are not medically supported by the retrieved evidence.
\end{tcolorbox}
\caption{Human annotation criteria.}
\label{fig:human_annotation_criteria}
\end{figure*}

\begin{table}[!t]
\centering
\small
\renewcommand{\arraystretch}{1.12}
\setlength{\tabcolsep}{6pt}
\caption{Human validation summary for sampled EHR-Complex tasks.}
\label{tab:human_validation}
\begin{tabular}{lcc}
\toprule
\textbf{Metric} & \textbf{Value} & \textbf{Count} \\
\midrule
Unsolved candidate tasks & 24.4\% & 954 / 3{,}915 \\
Audited sample & 52.4\% & 500 / 954 \\
Structurally unique templates & -- & 390 \\
\midrule
Adjudicated valid tasks & 95.0\% & 475 / 500 \\
Annotator 1 valid labels & 95.0\% & 475 / 500 \\
Annotator 2 valid labels & 95.2\% & 476 / 500 \\
\midrule
Raw annotator agreement & 93.0\% & 465 / 500 \\
Cohen's $\kappa$ & 0.25 & -- \\
PABAK & 0.86 & -- \\
\bottomrule
\end{tabular}
\end{table}

\section{Representative Population-Level Failure Cases}
\label{sec:appendix_cohort_failures}

This appendix provides compact snapshots of representative population-level failures.
These examples support the observation in Section~\ref{pag:pop_harder} that cohort queries are harder than patient-level retrieval: the SQL often executes successfully, but a local assumption can silently change the cohort before aggregation.

\newcommand{\cohortErrorBox}[1]{%
  \begingroup
  \setlength{\fboxsep}{3pt}%
  \noindent\colorbox{orange!18}{%
    \parbox{\dimexpr\linewidth-2\fboxsep\relax}{#1}%
  }%
  \endgroup
}

\tcbset{
  cohortcase/.style={
    enhanced,
    width=\linewidth,
    colback=white,
    colframe=black!25,
    boxrule=0.4pt,
    arc=2pt,
    left=5pt,
    right=5pt,
    top=5pt,
    bottom=5pt,
    before skip=0.8em,
    after skip=0.8em,
    fontupper=\small,
    coltitle=black,
    fonttitle=\bfseries,
    colbacktitle=black!5,
    boxed title style={
      colframe=black!25,
      colback=black!5,
      boxrule=0.3pt,
      arc=2pt
    }
  }
}

Orange highlights mark the local assumption that changes the cohort definition.
The three representative cases are shown in Tables~\ref{tab:cohort_case1}--\ref{tab:cohort_case3}.

\begin{table*}[!t]
\begin{tcolorbox}[cohortcase,title={Case 1: Per-patient temporal alignment}]
\begin{tabularx}{\linewidth}{@{}p{0.22\linewidth}X@{}}
\textbf{Question} &
What is the total number of individual patients whose first hospitalization had a diagnosis of Colonic diverticulosis (no bleed) and included lab test eGFR (MDRD), and who then, on their next (second) inpatient encounter, had any microbiology testing performed? \\
\addlinespace[0.4ex]

\textbf{Agent assumption} &
\cohortErrorBox{The agent first finds the earliest \emph{qualifying} admission for each patient and then uses \texttt{admission\_order = f.admission\_order + 1} to define the next admission.} \\
\addlinespace[0.4ex]

\textbf{Agent output} &
Final answer: 771 \\
\addlinespace[0.4ex]

\textbf{Ground truth} &
250 \\
\addlinespace[0.4ex]

\textbf{Analysis} &
The task requires the patient's \emph{first hospitalization overall} to satisfy the diagnosis and lab conditions. The agent instead anchors on the first qualifying admission anywhere in the timeline, which shifts the next-admission alignment and inflates the cohort count.
\end{tabularx}
\end{tcolorbox}
\caption{Representative population-level failure case: per-patient temporal alignment.}
\label{tab:cohort_case1}
\end{table*}

\begin{table*}[!t]
\begin{tcolorbox}[cohortcase,title={Case 2: Silent aggregate failure}]
\begin{tabularx}{\linewidth}{@{}p{0.22\linewidth}X@{}}
\textbf{Question} &
What are the top 5 most common insurance types among patients who had an index admission for ICD 43400 and a documented lab 51301 during a subsequent hospital admission? \\
\addlinespace[0.4ex]

\textbf{Agent assumption} &
\cohortErrorBox{The agent treats the subsequent admission as a 30-day bounce-back and adds \texttt{DATE\_DIFF('day', ia.index\_dischtime, a.admittime) <= 30}.} \\
\addlinespace[0.4ex]

\textbf{Agent output} &
Medicaid 2; Private 1. Final answer: Medicaid, Private \\
\addlinespace[0.4ex]

\textbf{Ground truth} &
Medicare 6, Medicaid 6, Private 2 \\
\addlinespace[0.4ex]

\textbf{Analysis} &
The benchmark asks for a subsequent admission, not a 30-day readmission. The added time window silently shrinks the cohort, while the query still returns a plausible insurance summary.
\end{tabularx}
\end{tcolorbox}
\caption{Representative population-level failure case: silent aggregate failure.}
\label{tab:cohort_case2}
\end{table*}

\begin{table*}[t!]
\begin{tcolorbox}[cohortcase,title={Case 3: Multi-stage query composition}]
\begin{tabularx}{\linewidth}{@{}p{0.22\linewidth}X@{}}
\textbf{Question} &
How many unique patients with an index admission diagnosis code \texttt{J45909} went on to have at least one later hospital admission that included both a FSH laboratory test and a prescription for Acetaminophen? \\
\addlinespace[0.4ex]

\textbf{Agent assumption} &
\cohortErrorBox{The agent broadens the target asthma diagnosis code to \texttt{493*}/\texttt{J45*} codes and matches prescriptions with \texttt{p.drug ILIKE '\%acetaminophen\%'}.} \\
\addlinespace[0.4ex]

\textbf{Agent output} &
unique\_patient\_count = 33. Final answer: 33 \\
\addlinespace[0.4ex]

\textbf{Ground truth} &
15 \\
\addlinespace[0.4ex]

\textbf{Analysis} &
The overall query structure is close, but two local relaxations change the cohort: the diagnosis condition is broader than the target asthma diagnosis code \texttt{J45909}, and the drug condition accepts substring matches rather than the exact drug string. Together, these choices increase the final count from 15 to 33.
\end{tabularx}
\end{tcolorbox}
\caption{Representative population-level failure case: multi-stage query composition.}
\label{tab:cohort_case3}
\end{table*}

\section{Error Analysis Details}
\label{sec:appendix_error_analysis}

\newcommand{\errspan}[1]{%
  \begingroup
  \setlength{\fboxsep}{2pt}%
  \noindent\colorbox{orange!12}{%
    \parbox{\dimexpr\linewidth-2\fboxsep\relax}{#1}%
  }%
  \endgroup
}

\lstdefinestyle{trajcode}{
  basicstyle=\ttfamily\small,
  breaklines=true,
  breakatwhitespace=false,
  columns=fullflexible,
  keepspaces=true,
  showstringspaces=false
}

\tcbset{
  trajcase/.style={
    enhanced,
    width=\linewidth,
    colback=white,
    colframe=black!22,
    boxrule=0.4pt,
    arc=2pt,
    left=5pt,
    right=5pt,
    top=5pt,
    bottom=5pt,
    before skip=0.7em,
    after skip=0.7em,
    fontupper=\small,
    coltitle=black,
    fonttitle=\bfseries,
    colbacktitle=black!5,
    boxed title style={
      colframe=black!22,
      colback=black!5,
      boxrule=0.3pt,
      arc=2pt
    }
  },
  trajcodebox/.style={
    enhanced,
    width=\linewidth,
    colback=black!1,
    colframe=black!18,
    boxrule=0.3pt,
    arc=2pt,
    left=5pt,
    right=5pt,
    top=4pt,
    bottom=4pt,
    before skip=0.5em,
    after skip=0.5em,
    fontupper=\small,
    coltitle=black,
    fonttitle=\bfseries,
    colbacktitle=black!4,
    boxed title style={
      colframe=black!18,
      colback=black!4,
      boxrule=0.3pt,
      arc=2pt
    }
  },
  protocolbox/.style={
    enhanced,
    breakable,
    width=\linewidth,
    colback=casebeige!45,
    colframe=casebeige!70!black,
    boxrule=0.4pt,
    arc=2pt,
    left=6pt,
    right=6pt,
    top=5pt,
    bottom=5pt,
    before skip=0.6em,
    after skip=0.6em,
    fontupper=\small,
    coltitle=black,
    fonttitle=\bfseries,
    colbacktitle=casebeige!65,
    boxed title style={
      colframe=casebeige!70!black,
      colback=casebeige!65,
      boxrule=0.3pt,
      arc=2pt
    }
  }
}

\subsection{Taxonomy Construction}
\label{sec:appendix_error_taxonomy}

We analyze failed trajectories from two representative models, Kimi-K2.5 and GPT-4.1, to characterize recurring failure modes on EHR-Complex.
Annotators first manually inspect more than 50 failures per model, reviewing the clinical question, intermediate SQL/Python actions, execution feedback, and final answer to identify the primary local error that causes divergence.
Based on this inspection, we define a compact taxonomy and use Gemini~2.5~Pro to annotate the remaining failures at scale under the same rubric, with ambiguous or low-confidence cases manually reviewed by annotators before aggregate statistics are reported.

\begin{table*}[t]
\centering
\small
\renewcommand{\arraystretch}{1.10}
\setlength{\tabcolsep}{4pt}
\begin{tabularx}{\linewidth}{@{}p{0.31\linewidth}X@{}}
\toprule
\textbf{Category} & \textbf{Definition} \\
\midrule
Semantic Misunderstanding
& Misinterprets the clinical request, such as the wrong admission, cohort criterion, or clinical ontology. \\
Medical-Code Lookup Error
& Fails to ground clinical terms to database-native identifiers, including ICD, laboratory, medication, or procedure codes. \\
SQL Logic Error
& Implements incorrect query logic, such as wrong joins, filters, aggregation keys, deduplication units, or temporal constraints. \\
Variable Scope Error
& Assumes variables or intermediate tables persist across isolated tool executions. \\
Answer Format Error
& Retrieves near-correct information but violates the required final-answer format. \\
Stuck-in-Loop
& Repeats unproductive variants of the same plan without meaningful revision. \\
\bottomrule
\end{tabularx}
\caption{Failure taxonomy used for trajectory annotation.}
\label{tab:error_taxonomy}
\end{table*}

\subsection{Comparative Failure Patterns}
\label{sec:appendix_error_patterns}

The two models exhibit different failure profiles.
Kimi-K2.5 is dominated by Semantic Misunderstanding, accounting for 38.0\% of its failures. These errors often arise before SQL construction and then propagate through otherwise executable queries.
GPT-4.1 is dominated by SQL Logic Error, accounting for 41.1\% of its failures. These cases often reflect incorrect joins, aggregation keys, temporal filters, or patient-level deduplication rules after the broad task intent has been identified.

Medical-Code Lookup Error is the second-largest category for both models, accounting for 28.6\% of Kimi-K2.5 failures and 32.1\% of GPT-4.1 failures.
This indicates that mapping clinical language to database-native identifiers remains a common bottleneck.
For example, MIMIC-IV stores ICD codes without dot notation, such as \texttt{25000} rather than \texttt{250.00}, which can differ from the surface forms models encounter in medical text~\citep{soroush2024large}.

Turn counts further distinguish the failure styles.
Kimi-K2.5 has long Stuck-in-Loop failures, averaging 44.7 turns, whereas GPT-4.1's Variable Scope Error cases are the most turn-intensive within its profile, averaging 17.1 turns.
These patterns suggest that execution feedback is only partially informative: the database may return clean outputs even when a local concept, code, join key, or temporal condition has already drifted from the intended clinical semantics.

\setcounter{dbltopnumber}{3}
\setcounter{topnumber}{5}
\setcounter{totalnumber}{8}

\renewcommand{\dbltopfraction}{0.98}
\renewcommand{\topfraction}{0.98}
\renewcommand{\textfraction}{0.01}
\renewcommand{\floatpagefraction}{0.80}
\renewcommand{\dblfloatpagefraction}{0.80}

\subsection{Representative Failure Trajectories}
\label{sec:appendix_trajectories}

The following examples illustrate the six error categories.
Each case shows the task snapshot, the key erroneous excerpt, and a brief analysis of how the local mistake propagates to the final answer.

\subsubsection{Trajectory 1: SQL Logic Error}

\noindent\textit{Corresponding case: Table~\ref{tab:traj1}.}

\begin{table*}[t]
\centering
\begin{tcolorbox}[trajcase,title={Trajectory 1: SQL Logic Error --- Task snapshot}]
\begin{tabularx}{\linewidth}{@{}p{0.22\linewidth}X@{}}
\textbf{Question} &
Among patients who underwent laboratory test 50820 during a hospital admission, which 5 ICD diagnosis codes were recorded for the largest number of distinct patients in the same admission? \\[0.5ex]

\textbf{Ground truth} &
\texttt{[E785, 4019, I10, Z87891, N179]} \\[0.5ex]

\textbf{Agent answer} &
\texttt{[D62, E785, J9601, 4019, E872]} \\[0.5ex]

\textbf{Local error} &
Counts joined lab--diagnosis rows rather than distinct patients. \\
\end{tabularx}
\end{tcolorbox}

\vspace{0.8ex}

\begin{tcolorbox}[trajcodebox,title={Trajectory 1: SQL Logic Error --- Key excerpt}]
\begin{lstlisting}[style=trajcode,language=SQL]
SELECT
    d.icd_code,
    d.icd_version,
    COUNT(*) AS frequency
FROM labevents l
JOIN diagnoses_icd d
    ON l.subject_id = d.subject_id
   AND l.hadm_id = d.hadm_id
WHERE l.itemid = 50820
GROUP BY d.icd_code, d.icd_version
ORDER BY frequency DESC
LIMIT 5
\end{lstlisting}

\errspan{\texttt{COUNT(*)} counts all joined lab--diagnosis combinations. The intended unit is distinct patients, e.g., \texttt{COUNT(DISTINCT d.subject\_id)}.}

\begin{lstlisting}[style=trajcode]
Agent output:
D62, E785, J9601, 4019, E872
\end{lstlisting}
\end{tcolorbox}

\caption[Trajectory 1: SQL Logic Error]{\textbf{Analysis.}
The query executes and returns plausible ICD codes, but the counting unit is wrong.
Because each admission can contain multiple lab events and diagnoses, counting joined rows instead of distinct patients inflates frequencies and changes the top-5 ranking.}
\label{tab:traj1}
\end{table*}

\subsubsection{Trajectory 2: Medical-Code Lookup Error}

\noindent\textit{Corresponding case: Table~\ref{tab:traj2}.}

\begin{table*}[t]
\centering
\begin{tcolorbox}[trajcase,title={Trajectory 2: Medical-Code Lookup Error --- Task snapshot}]
\begin{tabularx}{\linewidth}{@{}p{0.22\linewidth}X@{}}
\textbf{Question} &
Which five entry points contributed the most admissions for patients with ICD code Ileus who underwent laboratory test Vanc during their stay? \\[0.5ex]

\textbf{Ground truth} &
EMERGENCY ROOM; TRANSFER FROM HOSPITAL; PHYSICIAN REFERRAL; PACU; CLINIC REFERRAL \\[0.5ex]

\textbf{Agent answer} &
TRANSFER FROM HOSPITAL; PHYSICIAN REFERRAL; EMERGENCY ROOM; WALK-IN/SELF REFERRAL; TRANSFER FROM SKILLED NURSING FACILITY \\[0.5ex]

\textbf{Local error} &
After finding relevant Ileus codes, the agent hardcodes only a narrow ICD-10 subset and ignores dominant ICD-9 codes. \\
\end{tabularx}
\end{tcolorbox}

\vspace{0.8ex}

\begin{tcolorbox}[trajcodebox,title={Trajectory 2: Medical-Code Lookup Error --- Key excerpt}]
\begin{lstlisting}[style=trajcode]
ICD codes for Ileus:
icd_code  icd_version  long_title
27700     9            Cystic fibrosis w/o meconium ileus
27701     9            Cystic fibrosis with meconium ileus
5372      9            Chronic duodenal ileus
5601      9            Paralytic ileus
...       ...          ...
\end{lstlisting}

\errspan{The later SQL keeps only \texttt{('K560', 'K563', 'K567')} with \texttt{icd\_version = 10}, discarding ICD-9 codes such as \texttt{5601}.}

\begin{lstlisting}[style=trajcode,language=SQL]
WITH ileus_patients AS (
    SELECT DISTINCT subject_id, hadm_id
    FROM diagnoses_icd
    WHERE icd_code IN ('K560', 'K563', 'K567')
      AND icd_version = 10
)
\end{lstlisting}
\end{tcolorbox}

\caption[Trajectory 2: Medical-Code Lookup Error]{\textbf{Analysis.}
The agent performs an initially useful code lookup but fails to use the returned identifiers.
The resulting cohort is much smaller and has a different admission-location distribution.}
\label{tab:traj2}
\end{table*}

\subsubsection{Trajectory 3: Semantic Misunderstanding}

\noindent\textit{Corresponding case: Table~\ref{tab:traj3}.}

\begin{table*}[t]
\centering
\begin{tcolorbox}[trajcase,title={Trajectory 3: Semantic Misunderstanding --- Task snapshot}]
\begin{tabularx}{\linewidth}{@{}p{0.22\linewidth}X@{}}
\textbf{Question} &
What are the top 5 most common last care units during the second hospital admission for patients with an initial diagnosis of ICD code 486, given they had laboratory test 51221 during their first admission and laboratory test 50983 during their second? \\[0.5ex]

\textbf{Ground truth} &
Medicine; Hematology/Oncology; Medicine/Cardiology; Med/Surg; Neurology \\[0.5ex]

\textbf{Agent answer} &
Medical Intensive Care Unit (MICU); Medical/Surgical Intensive Care Unit (MICU/SICU); Surgical Intensive Care Unit (SICU); Coronary Care Unit (CCU); Trauma SICU (TSICU) \\[0.5ex]

\textbf{Local error} &
Interprets ``care unit'' as ICU-specific units and queries \texttt{icustays} instead of \texttt{transfers}. \\
\end{tabularx}
\end{tcolorbox}

\vspace{0.8ex}

\begin{tcolorbox}[trajcodebox,title={Trajectory 3: Semantic Misunderstanding --- Key excerpt}]
\begin{lstlisting}[style=trajcode,language=SQL]
SELECT
    i.last_careunit,
    COUNT(*) AS count
FROM icustays i
JOIN admissions_ordered ao ON i.hadm_id = ao.hadm_id
WHERE ao.admission_num = 2
  AND i.hadm_id IN (
      SELECT hadm_id FROM labevents WHERE itemid = 50983
  )
GROUP BY i.last_careunit
ORDER BY count DESC
LIMIT 5
\end{lstlisting}

\errspan{\texttt{icustays} contains ICU-specific units. The question asks for last care units during admission, which should be derived from \texttt{transfers}.}

\begin{lstlisting}[style=trajcode]
Agent output:
MICU, MICU/SICU, SICU, CCU, TSICU
\end{lstlisting}
\end{tcolorbox}

\caption[Trajectory 3: Semantic Misunderstanding]{\textbf{Analysis.}
The SQL is internally coherent but operates on the wrong clinical entity.
As a result, the answer is restricted to ICU units and misses the general ward units in the ground truth.}
\label{tab:traj3}
\end{table*}

\subsubsection{Trajectory 4: Variable Scope Error}

\noindent\textit{Corresponding case: Table~\ref{tab:traj4}.}

\begin{table*}[t]
\centering
\begin{tcolorbox}[trajcase,title={Trajectory 4: Variable Scope Error --- Task snapshot}]
\begin{tabularx}{\linewidth}{@{}p{0.22\linewidth}X@{}}
\textbf{Question} &
Across the cohort, what's the mean number of days from a patient's first admission coded Depression to their first subsequent readmission where Tbili was obtained? \\[0.5ex]

\textbf{Ground truth} &
\texttt{726.78 days} \\[0.5ex]

\textbf{Agent answer} &
\texttt{341.08 days} \\[0.5ex]

\textbf{Local error} &
Assumes variables persist across isolated code executions, repeatedly triggering \texttt{NameError} and later reconstructing a truncated cohort. \\
\end{tabularx}
\end{tcolorbox}

\vspace{0.8ex}

\begin{tcolorbox}[trajcodebox,title={Trajectory 4: Variable Scope Error --- Key excerpt}]
\begin{lstlisting}[style=trajcode]
Turn 1:
depression_codes = conn.execute("""
    SELECT icd_code, icd_version
    FROM d_icd_diagnoses
    WHERE LOWER(long_title) LIKE '%depression%'
""").df()

Turn 3:
NameError: name 'depression_codes' is not defined

Turn 5:
NameError: name 'depression_admissions' is not defined

Turn 7:
NameError: name 'pd' is not defined

Turn 9:
NameError: name 'readmissions' is not defined

Turn 11:
NameError: name 'tbili_lab_events' is not defined
\end{lstlisting}

\errspan{Repeated state-loss errors force redundant reconstruction. The final consolidated query produces 341.08 days instead of 726.78 days.}
\end{tcolorbox}

\caption[Trajectory 4: Variable Scope Error]{\textbf{Analysis.}
The environment is stateless across tool calls, but the agent repeatedly assumes previous variables remain available.
After several recovery attempts, the final query executes but uses an incomplete cohort.}
\label{tab:traj4}
\end{table*}

\subsubsection{Trajectory 5: Stuck-in-Loop}

\noindent\textit{Corresponding case: Table~\ref{tab:traj5}.}

\begin{table*}[t]
\centering
\begin{tcolorbox}[trajcase,title={Trajectory 5: Stuck-in-Loop --- Task snapshot}]
\begin{tabularx}{\linewidth}{@{}p{0.22\linewidth}X@{}}
\textbf{Question} &
During the hospitalization immediately following patient 10000980's initial stay with ICD-9 diagnosis code \texttt{496}, how many Chloride tests were ordered? \\[0.5ex]

\textbf{Ground truth} &
\texttt{7} \\[0.5ex]

\textbf{Agent answer} &
\texttt{None} \\[0.5ex]

\textbf{Local error} &
Interprets ICD-9 diagnosis code \texttt{496} as a literal \texttt{stay\_id}, then repeatedly searches for nonexistent stay identifiers. \\
\end{tabularx}
\end{tcolorbox}

\vspace{0.8ex}

\begin{tcolorbox}[trajcodebox,title={Trajectory 5: Stuck-in-Loop --- Key excerpt}]
\begin{lstlisting}[style=trajcode,language=SQL]
-- Initial mistaken hypothesis
SELECT * FROM icustays
WHERE subject_id = 10000980 AND stay_id = 496;

-- Empty result, but the agent keeps the same interpretation
SELECT * FROM icustays
WHERE CAST(stay_id AS VARCHAR) LIKE '%496%';

-- Later repeated through the remaining turns
SELECT * FROM icustays
WHERE subject_id = 10000980 AND stay_id = 30496496;
\end{lstlisting}

\errspan{The agent never revises the assumption that \texttt{496} is a stay identifier, even though the question specifies it as an ICD-9 diagnosis code.}
\end{tcolorbox}

\caption[Trajectory 5: Stuck-in-Loop]{\textbf{Analysis.}
The failure begins with a semantic misinterpretation of the diagnosis code as an ICU stay identifier, but the dominant behavior is lack of hypothesis revision.
Even after repeated empty results, the agent continues searching for guessed \texttt{stay\_id} values until the turn budget is exhausted.}
\label{tab:traj5}
\end{table*}

\subsubsection{Trajectory 6: Answer Format Error}

\noindent\textit{Corresponding case: Table~\ref{tab:traj6}.}

\begin{table*}[t]
\centering
\begin{tcolorbox}[trajcase,title={Trajectory 6: Answer Format Error --- Task snapshot}]
\begin{tabularx}{\linewidth}{@{}p{0.22\linewidth}Y@{}}
\textbf{Question} &
Looking across all admissions with 414.01 documented and a 50899 lab obtained, which five medications are most commonly ordered, based on the count of separate admissions in which each drug was prescribed? \\[0.5ex]

\textbf{Ground truth} &
Acetaminophen; Sodium Chloride 0.9\% Flush; Carbamazepine; Heparin; Docusate Sodium \\[0.5ex]

\textbf{Required format} &
A structured list with both \texttt{drug} and \texttt{prescription\_count}, e.g.,
\codefmt{{'drug': 'Acetaminophen', 'prescription_count': 31}}. \\[0.5ex]

\textbf{Agent answer} &
Acetaminophen; Sodium Chloride 0.9\% Flush; Carbamazepine; Heparin; Atorvastatin \\[0.5ex]

\textbf{Local error} &
Returns medication names without the required counts and does not follow the benchmark's deterministic answer-extraction outcome for the tied fifth medication. \\
\end{tabularx}
\end{tcolorbox}

\vspace{0.8ex}

\begin{tcolorbox}[trajcodebox,title={Trajectory 6: Answer Format Error --- Key excerpt}]
\begin{lstlisting}[style=trajcode,language=SQL]
SELECT p.drug, COUNT(DISTINCT p.hadm_id) AS admission_count
FROM prescriptions p
INNER JOIN common_admissions ca ON p.hadm_id = ca.hadm_id
GROUP BY p.drug
ORDER BY admission_count DESC
LIMIT 5
\end{lstlisting}

\begin{lstlisting}[style=trajcode]
Retrieved result:
Acetaminophen                    31
Sodium Chloride 0.9% Flush       29
Carbamazepine                    25
Heparin                          24
Atorvastatin                     23

Final answer:
Acetaminophen, Sodium Chloride 0.9% Flush,
Carbamazepine, Heparin, Atorvastatin
\end{lstlisting}

\errspan{The final answer omits the required count fields and does not match the benchmark's deterministic answer-extraction outcome for the tied fifth medication.}
\end{tcolorbox}

\caption[Trajectory 6: Answer Format Error]{\textbf{Analysis.}
The retrieval is close to correct, but the final response violates the required structured format by omitting \texttt{prescription\_count}.
It also returns \texttt{Atorvastatin} rather than the ground-truth tied medication \texttt{Docusate Sodium} under the benchmark's deterministic answer-extraction rule.
The error occurs at answer rendering rather than during SQL execution.}
\label{tab:traj6}
\end{table*}

\section{Pass\textasciicircum{}k and Pass@k Ablation Results}
\label{sec:appendix_passk}

This appendix presents detailed Pass\textasciicircum{}k and Pass@k results for the representative models used in the robustness analysis.
Pass\textasciicircum{}k (temperature=0.0) measures the proportion of tasks where \emph{all} k runs are correct, whereas Pass@k (temperature=0.6) measures the proportion of tasks where \emph{at least one} of k runs is correct.
Both metrics are estimated using the unbiased combinatorial estimator over n=4 independent runs.

\subsection{Overall Results}
Table~\ref{tab:passk_overall_full} reports the overall robustness results across all evaluated models.
The results show the expected contrast: Pass\textasciicircum{}k decreases as k increases, while Pass@k increases with additional samples.

\begin{table*}[t]
\centering
\scriptsize
\renewcommand{\arraystretch}{1.08}
\setlength{\tabcolsep}{3pt}
\resizebox{0.86\textwidth}{!}{%
\begin{tabular}{l|cccc|cccc}
\toprule
& \multicolumn{4}{c|}{\textbf{Pass\textasciicircum{}k (temp=0.0)}} & \multicolumn{4}{c}{\textbf{Pass@k (temp=0.6)}} \\
\textbf{Model} & k=1 & k=2 & k=3 & k=4 & k=1 & k=2 & k=3 & k=4 \\
\midrule
Qwen3.5-397B & 62.31 & 57.18 & 54.42 & 52.69 & 62.60 & 67.24 & 69.18 & 70.20 \\
Kimi-K2.5 & 60.74 & 54.96 & 51.87 & 49.81 & 59.90 & 66.50 & 69.35 & 71.14 \\
DeepSeek-V3.2-Exp & 58.23 & 50.89 & 46.35 & 42.95 & 58.83 & 66.43 & 69.45 & 71.17 \\
Qwen3-235B & 50.57 & 40.57 & 35.19 & 31.72 & 50.72 & 60.43 & 64.86 & 67.88 \\
GPT-4.1-mini & 47.12 & 37.55 & 32.49 & 29.33 & 45.85 & 55.26 & 59.55 & 62.31 \\
GPT-4.1 & 44.72 & 35.18 & 30.06 & 26.76 & 45.16 & 55.49 & 60.24 & 63.07 \\
Qwen3-32B & 34.33 & 23.53 & 18.36 & 15.35 & 33.63 & 46.15 & 53.31 & 58.31 \\
Gemini-2.5-Pro & 31.73 & 20.46 & 16.01 & 13.84 & 31.00 & 45.56 & 54.40 & 60.05 \\
GPT-4o & 28.71 & 18.80 & 14.44 & 12.18 & 28.59 & 38.79 & 44.30 & 47.86 \\
\bottomrule
\end{tabular}
}
\caption{Overall Pass\textasciicircum{}k and Pass@k results for k=1, 2, 3, and 4.}
\label{tab:passk_overall_full}
\end{table*}

\subsection{Results by Query Scope}
Table~\ref{tab:passk_by_scope} separates the robustness results by query scope.
Population-level tasks show substantially lower Pass\textasciicircum{}k consistency than patient-level tasks, matching the main-paper finding that cohort reasoning is more fragile.

\begin{table*}[t]
\centering
\scriptsize
\renewcommand{\arraystretch}{1.08}
\setlength{\tabcolsep}{2pt}
\begin{minipage}[t]{0.49\textwidth}
\centering
\textbf{(a) Patient-level tasks}\vspace{0.3ex}

\resizebox{\linewidth}{!}{%
\begin{tabular}{l|cccc|cccc}
\toprule
& \multicolumn{4}{c|}{\textbf{Pass\textasciicircum{}k}} & \multicolumn{4}{c}{\textbf{Pass@k}} \\
\textbf{Model} & k=1 & k=2 & k=3 & k=4 & k=1 & k=2 & k=3 & k=4 \\
\midrule
Qwen3.5-397B & 88.11 & 84.00 & 81.34 & 79.49 & 88.93 & 92.29 & 93.29 & 93.71 \\
Kimi-K2.5 & 87.75 & 83.57 & 81.27 & 79.77 & 86.85 & 92.19 & 94.18 & 95.40 \\
DeepSeek-V3.2-Exp & 83.62 & 76.58 & 71.48 & 67.53 & 82.50 & 89.95 & 92.07 & 93.14 \\
Qwen3-235B & 76.21 & 66.14 & 59.21 & 54.00 & 77.91 & 87.64 & 90.84 & 92.61 \\
GPT-4.1-mini & 69.87 & 58.83 & 52.56 & 48.43 & 69.11 & 80.30 & 84.73 & 87.50 \\
GPT-4.1 & 70.45 & 60.27 & 54.19 & 49.72 & 69.18 & 81.34 & 86.22 & 88.64 \\
Qwen3-32B & 50.58 & 36.15 & 28.48 & 23.84 & 48.26 & 65.46 & 75.00 & 81.70 \\
Gemini-2.5-Pro & 43.11 & 27.70 & 21.73 & 19.03 & 42.68 & 62.31 & 73.79 & 80.40 \\
GPT-4o & 46.95 & 33.00 & 26.42 & 23.01 & 47.30 & 62.12 & 69.89 & 74.72 \\
\bottomrule
\end{tabular}
}
\end{minipage}
\hfill
\begin{minipage}[t]{0.49\textwidth}
\centering
\textbf{(b) Population-level tasks}\vspace{0.3ex}

\resizebox{\linewidth}{!}{%
\begin{tabular}{l|cccc|cccc}
\toprule
& \multicolumn{4}{c|}{\textbf{Pass\textasciicircum{}k}} & \multicolumn{4}{c}{\textbf{Pass@k}} \\
\textbf{Model} & k=1 & k=2 & k=3 & k=4 & k=1 & k=2 & k=3 & k=4 \\
\midrule
Qwen3.5-397B & 41.20 & 35.24 & 32.40 & 30.77 & 41.26 & 46.95 & 49.65 & 51.16 \\
Kimi-K2.5 & 39.10 & 32.04 & 28.31 & 25.80 & 38.33 & 45.94 & 49.48 & 51.72 \\
DeepSeek-V3.2-Exp & 38.10 & 30.52 & 26.42 & 23.46 & 39.75 & 47.47 & 51.21 & 53.46 \\
Qwen3-235B & 29.94 & 20.00 & 15.86 & 13.79 & 29.07 & 38.76 & 44.17 & 48.19 \\
GPT-4.1-mini & 28.98 & 20.57 & 16.48 & 14.09 & 27.42 & 35.40 & 39.58 & 42.34 \\
GPT-4.1 & 24.32 & 15.28 & 10.92 & 8.56 & 26.13 & 34.98 & 39.64 & 42.79 \\
Qwen3-32B & 21.90 & 13.88 & 10.63 & 8.86 & 21.82 & 30.58 & 35.81 & 39.44 \\
Gemini-2.5-Pro & 22.69 & 14.71 & 11.46 & 9.71 & 21.73 & 32.28 & 39.02 & 43.92 \\
GPT-4o & 13.99 & 7.34 & 4.76 & 3.44 & 13.69 & 20.21 & 23.93 & 26.47 \\
\bottomrule
\end{tabular}
}
\end{minipage}
\caption{Pass\textasciicircum{}k and Pass@k results by query scope.}
\label{tab:passk_by_scope}
\label{tab:passk_single_patient}
\label{tab:passk_cohort_patient}
\end{table*}

\subsection{Results by Clinical Intent}
Tables~\ref{tab:passk_intents_cb_dt}--\ref{tab:passk_intents_mo_vo} report the intent-level k-sweeps.
Across intents, Cost is closest to saturation under sampling, whereas Diagnoses, Vitals, and Medications remain more brittle under repeated rollouts.

\begin{table*}[t]
\centering
\scriptsize
\renewcommand{\arraystretch}{1.08}
\setlength{\tabcolsep}{2pt}
\begin{minipage}[t]{0.49\textwidth}
\centering
\textbf{(a) Cost \& Billing}\vspace{0.3ex}

\resizebox{\linewidth}{!}{%
\begin{tabular}{l|cccc|cccc}
\toprule
& \multicolumn{4}{c|}{\textbf{Pass\textasciicircum{}k}} & \multicolumn{4}{c}{\textbf{Pass@k}} \\
\textbf{Model} & k=1 & k=2 & k=3 & k=4 & k=1 & k=2 & k=3 & k=4 \\
\midrule
Qwen3.5-397B & 84.62 & 82.05 & 80.77 & 80.77 & 85.58 & 87.82 & 88.46 & 88.46 \\
Kimi-K2.5 & 86.54 & 82.05 & 78.85 & 76.92 & 80.77 & 84.62 & 86.54 & 88.46 \\
DeepSeek-V3.2-Exp & 82.69 & 77.56 & 73.08 & 69.23 & 86.54 & 88.46 & 88.46 & 88.46 \\
Qwen3-235B & 83.65 & 79.49 & 75.96 & 73.08 & 82.69 & 88.46 & 88.46 & 88.46 \\
GPT-4.1-mini & 84.62 & 81.41 & 78.85 & 76.92 & 81.73 & 88.46 & 91.35 & 92.31 \\
GPT-4.1 & 75.00 & 69.23 & 65.38 & 61.54 & 76.92 & 86.54 & 90.38 & 92.31 \\
Qwen3-32B & 64.42 & 53.21 & 47.12 & 42.31 & 73.96 & 83.33 & 89.58 & 95.83 \\
Gemini-2.5-Pro & 50.00 & 36.54 & 29.81 & 26.92 & 50.00 & 69.23 & 79.81 & 84.62 \\
GPT-4o & 58.65 & 47.44 & 40.38 & 34.62 & 63.00 & 78.67 & 82.00 & 84.00 \\
\bottomrule
\end{tabular}
}
\end{minipage}
\hfill
\begin{minipage}[t]{0.49\textwidth}
\centering
\textbf{(b) Demographics \& Tracking}\vspace{0.3ex}

\resizebox{\linewidth}{!}{%
\begin{tabular}{l|cccc|cccc}
\toprule
& \multicolumn{4}{c|}{\textbf{Pass\textasciicircum{}k}} & \multicolumn{4}{c}{\textbf{Pass@k}} \\
\textbf{Model} & k=1 & k=2 & k=3 & k=4 & k=1 & k=2 & k=3 & k=4 \\
\midrule
Qwen3.5-397B & 59.55 & 54.97 & 52.62 & 50.79 & 60.16 & 64.67 & 66.93 & 68.23 \\
Kimi-K2.5 & 63.02 & 57.56 & 54.64 & 52.58 & 60.88 & 67.27 & 69.69 & 70.98 \\
DeepSeek-V3.2-Exp & 57.60 & 50.60 & 46.65 & 43.81 & 59.28 & 67.18 & 70.62 & 72.68 \\
Qwen3-235B & 48.59 & 38.80 & 33.59 & 30.26 & 48.86 & 58.12 & 62.56 & 65.48 \\
GPT-4.1-mini & 50.26 & 40.85 & 35.64 & 32.31 & 48.10 & 56.68 & 60.53 & 62.94 \\
GPT-4.1 & 48.48 & 39.26 & 33.88 & 30.46 & 47.59 & 57.95 & 62.69 & 65.48 \\
Qwen3-32B & 32.21 & 22.16 & 17.35 & 14.12 & 30.80 & 42.82 & 49.59 & 54.14 \\
Gemini-2.5-Pro & 32.49 & 21.57 & 17.64 & 15.23 & 31.09 & 43.99 & 51.52 & 56.35 \\
GPT-4o & 28.97 & 18.55 & 13.59 & 10.26 & 28.30 & 38.24 & 44.42 & 48.73 \\
\bottomrule
\end{tabular}
}
\end{minipage}
\caption{Intent-level Pass\textasciicircum{}k and Pass@k results for Cost and Demographics tasks.}
\label{tab:passk_intents_cb_dt}
\label{tab:passk_cost___billing}
\label{tab:passk_demographics___tracking}
\end{table*}

\begin{table*}[t]
\centering
\scriptsize
\renewcommand{\arraystretch}{1.08}
\setlength{\tabcolsep}{2pt}
\begin{minipage}[t]{0.49\textwidth}
\centering
\textbf{(a) Diagnoses \& Procedures}\vspace{0.3ex}

\resizebox{\linewidth}{!}{%
\begin{tabular}{l|cccc|cccc}
\toprule
& \multicolumn{4}{c|}{\textbf{Pass\textasciicircum{}k}} & \multicolumn{4}{c}{\textbf{Pass@k}} \\
\textbf{Model} & k=1 & k=2 & k=3 & k=4 & k=1 & k=2 & k=3 & k=4 \\
\midrule
Qwen3.5-397B & 50.76 & 43.64 & 40.08 & 38.17 & 52.46 & 57.32 & 59.47 & 60.61 \\
Kimi-K2.5 & 51.88 & 45.11 & 41.35 & 39.10 & 50.95 & 58.96 & 62.31 & 64.39 \\
DeepSeek-V3.2-Exp & 48.87 & 39.60 & 34.02 & 30.08 & 52.71 & 61.63 & 64.92 & 66.67 \\
Qwen3-235B & 37.97 & 27.19 & 22.37 & 19.55 & 36.57 & 46.14 & 51.31 & 55.22 \\
GPT-4.1-mini & 34.02 & 24.06 & 18.98 & 15.79 & 34.70 & 44.53 & 49.44 & 52.24 \\
GPT-4.1 & 34.89 & 27.11 & 23.32 & 20.90 & 34.89 & 42.91 & 46.46 & 48.51 \\
Qwen3-32B & 24.79 & 16.24 & 12.18 & 10.26 & 25.43 & 36.06 & 42.46 & 46.55 \\
Gemini-2.5-Pro & 26.31 & 15.42 & 10.45 & 7.46 & 25.56 & 40.67 & 50.56 & 57.46 \\
GPT-4o & 21.35 & 14.10 & 10.77 & 9.23 & 18.61 & 25.94 & 30.45 & 33.08 \\
\bottomrule
\end{tabular}
}
\end{minipage}
\hfill
\begin{minipage}[t]{0.49\textwidth}
\centering
\textbf{(b) Labs \& Microbiology}\vspace{0.3ex}

\resizebox{\linewidth}{!}{%
\begin{tabular}{l|cccc|cccc}
\toprule
& \multicolumn{4}{c|}{\textbf{Pass\textasciicircum{}k}} & \multicolumn{4}{c}{\textbf{Pass@k}} \\
\textbf{Model} & k=1 & k=2 & k=3 & k=4 & k=1 & k=2 & k=3 & k=4 \\
\midrule
Qwen3.5-397B & 67.37 & 63.31 & 61.41 & 60.40 & 67.31 & 71.35 & 73.16 & 74.25 \\
Kimi-K2.5 & 64.07 & 59.27 & 56.62 & 54.64 & 64.35 & 70.47 & 73.07 & 74.50 \\
DeepSeek-V3.2-Exp & 63.29 & 56.70 & 52.33 & 48.84 & 63.25 & 70.03 & 72.43 & 73.84 \\
Qwen3-235B & 56.50 & 46.85 & 41.05 & 37.16 & 56.62 & 66.50 & 70.61 & 73.51 \\
GPT-4.1-mini & 49.92 & 39.90 & 34.69 & 31.46 & 48.18 & 58.03 & 62.05 & 64.69 \\
GPT-4.1 & 48.84 & 38.45 & 32.59 & 28.71 & 48.51 & 58.86 & 63.53 & 66.34 \\
Qwen3-32B & 36.36 & 24.12 & 18.18 & 14.77 & 35.04 & 48.82 & 56.88 & 62.83 \\
Gemini-2.5-Pro & 33.99 & 23.10 & 18.65 & 16.50 & 32.34 & 47.36 & 56.44 & 62.05 \\
GPT-4o & 31.60 & 21.40 & 16.91 & 14.85 & 32.34 & 43.78 & 49.42 & 52.81 \\
\bottomrule
\end{tabular}
}
\end{minipage}
\caption{Intent-level Pass\textasciicircum{}k and Pass@k results for Diagnoses and Labs tasks.}
\label{tab:passk_intents_dp_lm}
\label{tab:passk_diagnoses___procedures}
\label{tab:passk_labs___microbiology}
\end{table*}

\begin{table*}[t]
\centering
\scriptsize
\renewcommand{\arraystretch}{1.08}
\setlength{\tabcolsep}{2pt}
\begin{minipage}[t]{0.49\textwidth}
\centering
\textbf{(a) Medications \& Orders}\vspace{0.3ex}

\resizebox{\linewidth}{!}{%
\begin{tabular}{l|cccc|cccc}
\toprule
& \multicolumn{4}{c|}{\textbf{Pass\textasciicircum{}k}} & \multicolumn{4}{c}{\textbf{Pass@k}} \\
\textbf{Model} & k=1 & k=2 & k=3 & k=4 & k=1 & k=2 & k=3 & k=4 \\
\midrule
Qwen3.5-397B & 63.48 & 56.67 & 51.96 & 48.70 & 62.83 & 69.57 & 71.52 & 72.17 \\
Kimi-K2.5 & 54.57 & 47.10 & 43.70 & 41.74 & 54.96 & 62.93 & 67.24 & 70.69 \\
DeepSeek-V3.2-Exp & 53.73 & 44.88 & 39.69 & 35.96 & 49.12 & 58.19 & 62.50 & 64.91 \\
Qwen3-235B & 47.41 & 36.06 & 30.39 & 26.72 & 49.35 & 59.77 & 64.87 & 68.10 \\
GPT-4.1-mini & 43.97 & 33.91 & 28.45 & 25.00 & 41.67 & 51.00 & 55.98 & 59.83 \\
GPT-4.1 & 33.12 & 21.94 & 17.09 & 14.53 & 38.89 & 52.14 & 58.55 & 62.39 \\
Qwen3-32B & 34.86 & 24.36 & 19.71 & 17.31 & 34.05 & 46.98 & 53.33 & 57.14 \\
Gemini-2.5-Pro & 29.09 & 16.81 & 12.28 & 11.21 & 31.41 & 46.01 & 54.49 & 59.83 \\
GPT-4o & 21.96 & 11.88 & 8.26 & 6.96 & 23.72 & 34.05 & 40.17 & 44.44 \\
\bottomrule
\end{tabular}
}
\end{minipage}
\hfill
\begin{minipage}[t]{0.49\textwidth}
\centering
\textbf{(b) Vital Signs \& Observations}\vspace{0.3ex}

\resizebox{\linewidth}{!}{%
\begin{tabular}{l|cccc|cccc}
\toprule
& \multicolumn{4}{c|}{\textbf{Pass\textasciicircum{}k}} & \multicolumn{4}{c}{\textbf{Pass@k}} \\
\textbf{Model} & k=1 & k=2 & k=3 & k=4 & k=1 & k=2 & k=3 & k=4 \\
\midrule
Qwen3.5-397B & 52.63 & 45.61 & 40.79 & 36.84 & 50.00 & 54.63 & 55.56 & 55.56 \\
Kimi-K2.5 & 48.68 & 39.47 & 34.21 & 31.58 & 43.06 & 44.44 & 44.44 & 44.44 \\
DeepSeek-V3.2-Exp & 43.42 & 40.35 & 38.16 & 36.84 & 46.05 & 53.51 & 56.58 & 57.89 \\
Qwen3-235B & 40.79 & 28.95 & 23.68 & 21.05 & 40.79 & 54.39 & 60.53 & 63.16 \\
GPT-4.1-mini & 30.26 & 22.81 & 21.05 & 21.05 & 40.79 & 52.63 & 59.21 & 63.16 \\
GPT-4.1 & 39.47 & 32.46 & 28.95 & 26.32 & 34.21 & 42.98 & 48.68 & 52.63 \\
Qwen3-32B & 40.62 & 28.12 & 21.88 & 18.75 & 38.33 & 51.11 & 60.00 & 66.67 \\
Gemini-2.5-Pro & 17.11 & 2.63 & 0.00 & 0.00 & 18.42 & 32.46 & 43.42 & 52.63 \\
GPT-4o & 30.26 & 14.91 & 10.53 & 10.53 & 26.32 & 31.58 & 34.21 & 36.84 \\
\bottomrule
\end{tabular}
}
\end{minipage}
\caption{Intent-level Pass\textasciicircum{}k and Pass@k results for Medications and Vitals tasks.}
\label{tab:passk_intents_mo_vo}
\label{tab:passk_medications___orders}
\label{tab:passk_vital_signs___observations}
\end{table*}

\section{Interaction Budget Results}
\label{sec:appendix_budget}

This appendix reports full interaction-budget curves for the evaluated models.
SR(B) denotes the fraction of tasks solved within budget B, where B=1, 3, 5, 10, 15, and 20.
The final two columns report the full success rate under the original 50-turn cap and the mean number of interaction turns over all tasks.

Table~\ref{tab:interaction_budget_appendix} shows that the representative models used in the main interaction analysis are not outliers.
For most systems, success rises once several rounds of execution feedback are available and then saturates well before the 50-turn cap.

\begin{table*}[t]
\centering
\scriptsize
\renewcommand{\arraystretch}{1.08}
\setlength{\tabcolsep}{3pt}
\resizebox{0.90\textwidth}{!}{%
\begin{tabular}{l|cccccc|cc}
\toprule
\textbf{Model} & B=1 & B=3 & B=5 & B=10 & B=15 & B=20 & Full SR & Avg.\ turns \\
\midrule
Qwen3.5-397B & 0.00 & 0.00 & 0.05 & 0.39 & 0.54 & 0.58 & 0.60 & 11.82 \\
Kimi-K2.5 & 0.00 & 0.00 & 0.05 & 0.40 & 0.54 & 0.57 & 0.59 & 11.10 \\
DeepSeek-V3.2-Exp & 0.00 & 0.15 & 0.25 & 0.43 & 0.51 & 0.54 & 0.57 & 9.73 \\
DeepSeek-V3.1 & 0.00 & 0.02 & 0.20 & 0.49 & 0.52 & 0.53 & 0.53 & 7.41 \\
Qwen3-235B & 0.00 & 0.02 & 0.13 & 0.41 & 0.48 & 0.50 & 0.50 & 9.14 \\
GPT-4.1-mini & 0.00 & 0.10 & 0.17 & 0.34 & 0.42 & 0.44 & 0.45 & 8.85 \\
GPT-4.1 & 0.00 & 0.03 & 0.10 & 0.34 & 0.42 & 0.43 & 0.44 & 9.29 \\
Qwen3-32B & 0.00 & 0.14 & 0.24 & 0.32 & 0.33 & 0.33 & 0.33 & 5.35 \\
Gemini 2.5 Pro & 0.06 & 0.20 & 0.26 & 0.29 & 0.30 & 0.30 & 0.30 & 2.35 \\
Qwen3-14B & 0.00 & 0.12 & 0.19 & 0.25 & 0.27 & 0.27 & 0.27 & 6.46 \\
GPT-4o & 0.00 & 0.00 & 0.05 & 0.24 & 0.27 & 0.27 & 0.27 & 8.24 \\
\bottomrule
\end{tabular}
}
\caption{Interaction-budget results.}
\label{tab:interaction_budget_appendix}
\end{table*}

\section{SFT Training Details}
\label{sec:appendix_sft}

\subsection{Trajectory Data}
\label{sec:appendix_sft_data}

We extract 6{,}000 successful task completions generated by Kimi-K2.5 from the EHR-Complex training partition.
All selected trajectories are disjoint from the test set and have final answers that exactly match the verified ground truth.
Each trajectory contains the original clinical question, multi-turn SQL/Python execution attempts, database feedback, iterative query refinements, and the final verified answer.
The average trajectory contains 20.9 messages, preserving the interactive reasoning process rather than reducing supervision to static input--output pairs.

\subsection{Training Configuration}
\label{sec:appendix_sft_config}

Table~\ref{tab:sft_config} summarizes the supervised fine-tuning setup.
We use the same EHR-Complex test set and exact-match evaluation protocol as in the main experiments.

\begin{table*}[t]
\centering
\small
\renewcommand{\arraystretch}{1.12}
\setlength{\tabcolsep}{5pt}
\begin{tabularx}{0.92\textwidth}{@{}p{0.24\linewidth}X@{}}
\toprule
\textbf{Component} & \textbf{Setting} \\
\midrule
Training data
& 6{,}000 successful Kimi-K2.5 trajectories from the training set. All trajectories are test-disjoint and exact-match correct. \\

Trajectory format
& Original clinical question, multi-turn SQL/Python attempts, execution feedback, iterative query refinements, and final verified answer. Average length: 20.9 messages. \\

Base models
& Qwen3-14B and Qwen3-32B~\citep{qwen3}. \\

Optimization
& Full-parameter supervised fine-tuning for one epoch. \\

Distributed setup
& 8 nodes with 8$\times$H800 GPUs per node, for 64 GPUs total; DeepSpeed ZeRO-3. \\

Precision and context length
& bfloat16 precision; maximum sequence length 32{,}768 tokens. \\

Learning rate
& $1\times10^{-5}$ with 5\% warmup. \\

Effective batch size
& 128. \\

Selected checkpoints
& checkpoint-110 for Qwen3-14B; checkpoint-55 for Qwen3-32B. \\

Evaluation
& Same EHR-Complex test set and exact-match execution metric as Table~\ref{tab:main_results}. \\
\bottomrule
\end{tabularx}
\caption{Supervised fine-tuning configuration.}
\label{tab:sft_config}
\end{table*}

\subsection{Training Dynamics}
\label{sec:appendix_sft_dynamics}

\begin{figure*}[t]
  \centering
  \includegraphics[width=0.92\linewidth]{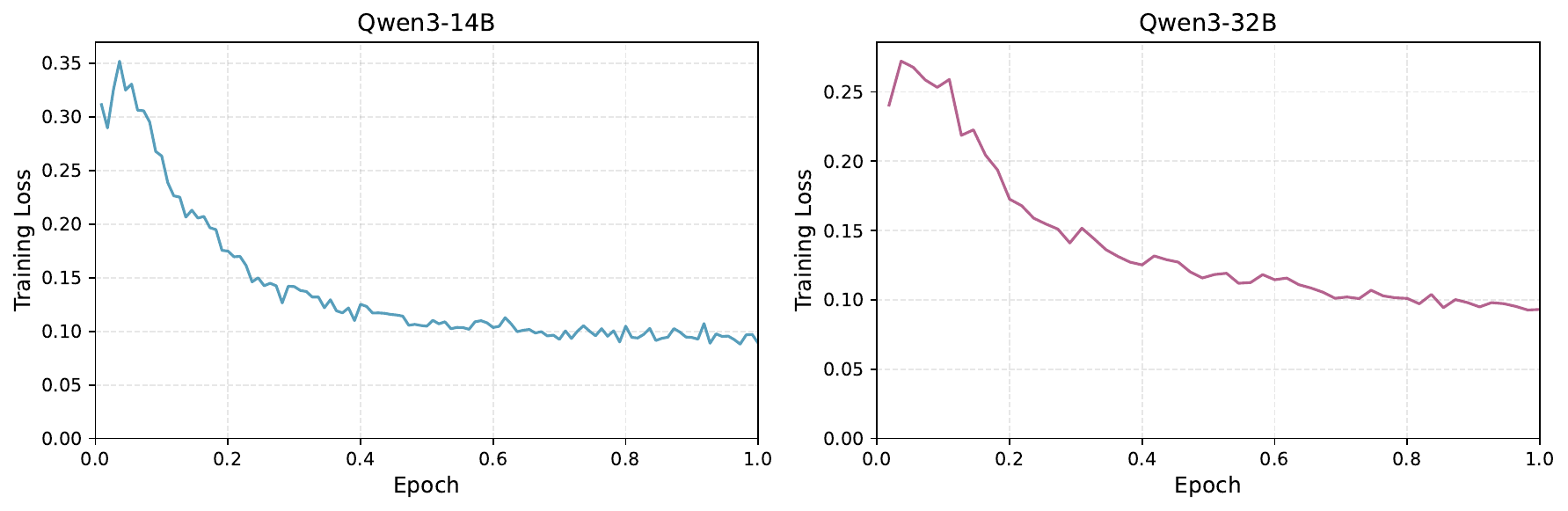}
  \caption{Training loss curves during supervised fine-tuning.}
  \label{fig:appendix_training_curves}
\end{figure*}

Figure~\ref{fig:appendix_training_curves} shows the training loss curves for the two fine-tuned models.
Both models fit the trajectory data within the one-epoch SFT setting.

The Qwen3-14B loss decreases from 0.311 to 0.089 over 110 steps, while the Qwen3-32B loss decreases from 0.240 to 0.093 over 55 steps.
These curves indicate that the models optimize quickly on the successful interaction trajectories.

\subsection{Additional SFT Result Analysis}
\label{sec:appendix_sft_results}

Both fine-tuned models improve across most clinical intents and query scopes.
Qwen3-14B-SFT reaches 0.45 macro-averaged accuracy, an absolute gain of 0.15 over the base model.
Qwen3-32B-SFT reaches 0.55, an absolute gain of 0.19 over the base model.
The largest gains appear on patient-level tasks requiring structured temporal or schema reasoning, including Labs Pat. for Qwen3-14B (0.40 $\rightarrow$ 0.74), Cost Pat. for Qwen3-14B (0.63 $\rightarrow$ 0.84), Demographics Pat. for Qwen3-32B (0.51 $\rightarrow$ 0.83), and Labs Pat. for Qwen3-32B (0.50 $\rightarrow$ 0.80).

These results should be interpreted as an in-domain training-signal validation.
The experiment uses a single teacher model and evaluates on the same benchmark distribution, so it does not establish transfer to other EHR databases or clinical environments.
Nevertheless, the gains suggest that successful EHR-Complex trajectories provide useful supervision for schema navigation, temporal reasoning, medical grounding, and execution-guided query refinement.

\end{document}